\documentclass[lettersize,journal]{IEEEtran}
\usepackage{amsmath,amsfonts}
\usepackage{algorithmic}
\usepackage{algorithm}
\usepackage{array}
\usepackage[caption=false,font=normalsize,labelfont=sf,textfont=sf]{subfig}
\usepackage{textcomp}
\usepackage{stfloats}
\usepackage{url}
\usepackage{verbatim}
\usepackage{graphicx}
\usepackage{xcolor}
\usepackage{cite}
\usepackage[numbers,sort&compress]{natbib}
\usepackage{booktabs} 
\usepackage{multirow}
\usepackage{cleveref}
\usepackage{mathrsfs}
\usepackage{mdframed}
\usepackage{amssymb}
\usepackage{natbib}
\usepackage{marvosym}
\usepackage{soul}

\begin{document}

\title{Instruct-ReID++: Towards Universal Purpose Instruction-Guided Person Re-identification}

\author{Weizhen He*, Yiheng Deng*, Yunfeng Yan\textsuperscript{\Letter}, Feng Zhu,
Yizhou Wang, Lei Bai, \\ Qingsong Xie, Rui Zhao, Donglian Qi, Wanli Ouyang,~\IEEEmembership{Senior Member,~IEEE}, Shixiang Tang\textsuperscript{\Letter}
        % <-this % stops a space
\thanks{$^{*}$Weizhen He and Yiheng Deng contributed equally. $\textsuperscript{\Letter}$Shixiang Tang and $\textsuperscript{\Letter}$ Yunfeng Yan are corresponding authors. This work was done when Weizhen He was an intern at SenseTime. Weizhen He, Yiheng Deng, Donglian Qi and Yunfeng Yan are with the College of Electrical Engineering, Zhejiang University, Hangzhou, 310027, China. Feng Zhu and Rui Zhao is with SenseTime Group Limited, China. Shixiang Tang, Yizhou Wang and Wanli~Ouyang are with Shanghai AI Laboratory, Shanghai, 200232, China. Qingsong Xie is with Shanghai Jiao Tong University, Shanghai, 200240, China.}% <-this % stops a space
\thanks{Manuscript received April 19, 2021; revised August 16, 2021.}}

\markboth{Journal of \LaTeX\ Class Files,~Vol.~14, No.~8, August~2021}%
{Shell \MakeLowercase{\textit{et al.}}: A Sample Article Using IEEEtran.cls for IEEE Journals}


\maketitle
% \renewcommand{\thefootnote}{\Letter}
% \footnotetext{Corresponding author.}
% \renewcommand{\thefootnote}{1}

\begin{abstract}
\textcolor{black}{Recently, person re-identification (ReID) has witnessed fast development due to its broad practical applications and proposed various settings, \emph{e.g.,} traditional ReID, clothes-changing ReID, and visible-infrared ReID. However, current studies primarily focus on single specific tasks, which limits model applicability in real-world scenarios. This paper aims to address this issue by introducing a novel instruct-ReID task that unifies \textbf{6} existing ReID tasks in one model and retrieves images based on provided visual or textual instructions. Instruct-ReID is the first exploration of a general ReID setting, where \textbf{6} existing ReID tasks can be viewed as special cases by assigning different instructions.} To facilitate research in this new instruct-ReID task, we propose a large-scale OmniReID++ benchmark equipped with diverse data and comprehensive evaluation methods, \emph{e.g.,} task-specific and task-free evaluation settings. In the task-specific evaluation setting, gallery sets are categorized according to specific ReID tasks. We propose a novel baseline model, IRM, with an adaptive triplet loss to handle various retrieval tasks within a unified framework. For task-free evaluation setting, where target person images are retrieved from task-agnostic gallery sets, we further propose a new method called IRM++ with novel memory bank-assisted learning. 
Extensive evaluations of IRM and IRM++ on OmniReID++ benchmark demonstrate the superiority of our proposed methods, achieving state-of-the-art performance on 10 test sets. The datasets, the model, and the code will be available at https://github.com/hwz-zju/Instruct-ReID.
\end{abstract}

\begin{IEEEkeywords}
Person Re-identification, Multitask Person Retrieval, Benchmark, General Foundation Model
\end{IEEEkeywords}

\section{Introduction}
\IEEEPARstart{P}ERSON re-identification (ReID) aims to retrieve the target person from surveillance videos or images across locations and time. This task is challenging due to various viewpoints, illumination changes, unconstrained poses, occlusions, heterogeneous modalities, background clutter, and more. Therefore, recent studies in person re-identification carefully designed task settings and developed state-of-the-art models to tackle every specific scenario, such as ~\cite{zheng2015scalable,chen2017person,wei2018person,Zheng_2021_ICCV,li2019pose} for traditional ReID (Trad-ReID), ~\cite{huang2021clothing,gu2022clothes,jin2022cloth,shu2021semantic,hong2021fine} for clothes-changing ReID (CC-ReID), ~\cite{yu2020cocas,li2022cocas+} for clothes template based clothes-changing ReID (CTCC-ReID), ~\cite{liu2022learning,yang2022augmented,yang2022learning,zhang2022fmcnet,10319076} for visible-infrared ReID (VI-ReID) and ~\cite{bai2023rasa,chen2018improving,zheng2020dual,li2017person,yin2017adversarial} for text-to-image ReID (T2I-ReID). 

Despite the success of these person re-identification methods, they still have inherent limitations in practical applications due to their focus on specific and fixed scenarios. First, customers must deploy distinct models to retrieve persons according to the person retrieval tasks, \emph{e.g.,} 5 models for 5 re-identification settings, significantly increasing the model training and deployment cost. Second, focusing on a single scenario task limits the model performance. ReID models utilize biological appearance cues to retrieve images with the same identity as the query image. A model trained across diverse task scenarios is expected to have a deeper understanding of person re-identification, \emph{e.g.,} the traditional ReID task requires the model to possess the ability to grasp characteristics such as body posture and appearance to compare identities, and clothes-changing ReID can assist the model in extracting non-clothing information for identity determination. Other tasks, \emph{e.g.,} text-to-image ReID and visible-infrared ReID, also help the model learn to extract human identity information across different data modalities. The methods designed for specific scenarios are limited to their particular training data and ignore the benefits of other ReID tasks, leading to limited model performance.
% although various ReID methods focus on identifying person identities across different re-identification settings, they fail to leverage the benefits provided by other ReID tasks, leading to a lower upper bound on performance. 
% Concretely, traditional ReID relies on diverse human image characteristics such as body posture and appearance to compare identities, while clothes-changing ReID requires extracting non-clothing information for identity determination. However, both ReID tasks use biological appearance cues to match images with the same identity as the query image. Similarly, clothes template clothes-changing ReID, text-to-image ReID, and visible-infrared ReID tasks all require the model to extract human identity information across different data modalities. Therefore, a model trained across multiple tasks is expected to have a deeper understanding of person retrieval, thereby enhancing its performance across various tasks.

\begin{figure*}
  \centering
  \includegraphics[width=\linewidth]
  {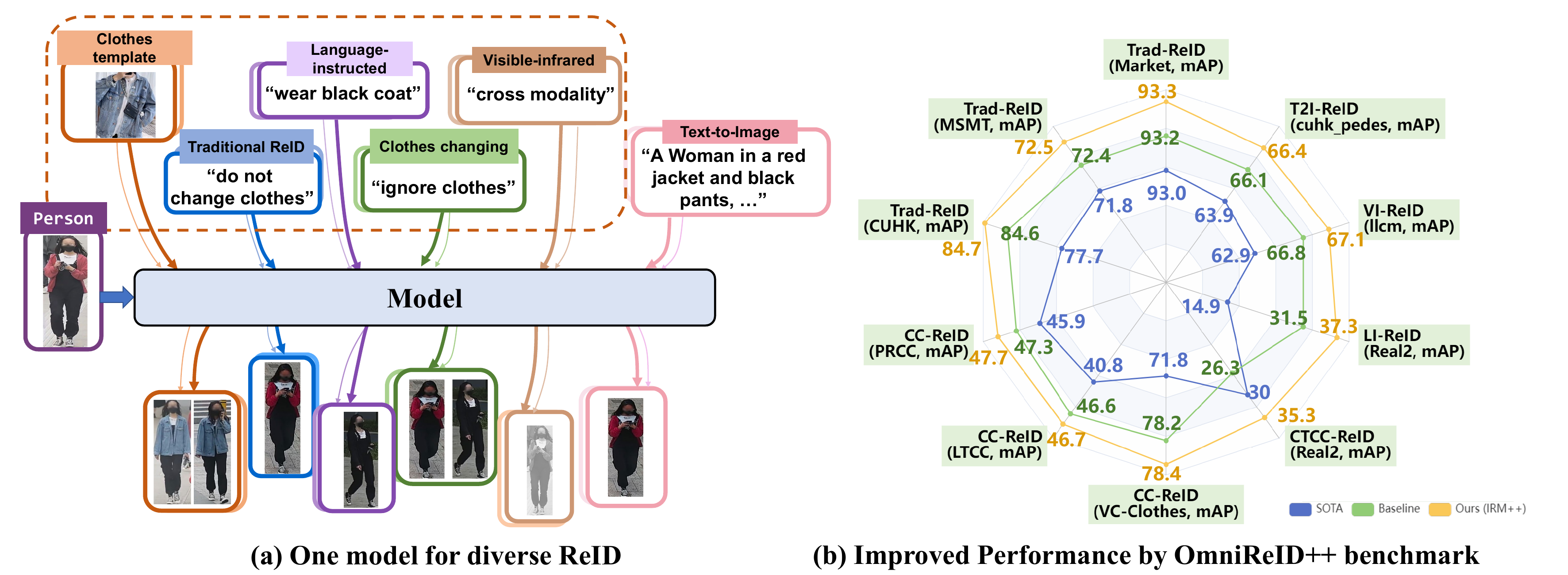}
  \caption{\textcolor{black}{(a)} We proposed a new instruct-ReID task that unites various ReID tasks. \emph{Traditional ReID:} The instruction may be “Do not change clothes". \emph{Clothes-changing ReID:} The instruction may be “Ignore clothes". \emph{Clothes template based clothes-changing ReID:} The instruction is a cropped clothes image and the model should retrieve the same person wearing the provided clothing. \emph{Language-instructed ReID:} The instruction is several sentences describing pedestrian attributes. The model is required to retrieve the person described by the instruction. \emph{Visible-Infrared ReID:} The instruction can be “Cross modality". \emph{Text-to-image ReID:} The model retrieves images according to the description sentence.
\textcolor{black}{(b) Our proposed method advances the performance limits of various person ReID tasks through a unified retrieval model. Specifically, on 10 datasets across the 6 ReID tasks, our method achieves a performance improvement of +0.3\% mAP to +22.4\% mAP compared to existing state-of-the-art methods}}
  \label{fig:overview}
\end{figure*}

% Introduce Instruct-ReID task
The limitations of addressing each ReID task independently motivate us to design a general approach towards distinct ReID tasks. In this paper, we propose a novel multi-purpose \textit{instruct-ReID} task where  \textbf{6} existing ReID settings, \emph{i.e.,} Trad-ReID, CC-ReID, CTCC-ReID, VI-ReID, T2I-ReID, and language-instructed ReID (LI-ReID) can be formulated as its special cases. This unified ReID task involves training various ReID tasks using a single model, thereby minimizing training costs and fostering mutual benefits among diverse ReID tasks. Specifically, the instruct-ReID task takes query images and multimodality instructions as the model inputs and requires the model to retrieve the same identity images from the gallery following the instructions. As shown in \textcolor{black}{Fig.~\ref{fig:overview} (a)}, with certain language or image instructions, the instruct-ReID task can be specialized to existing ReID tasks. For example, the clothes-changing ReID can be viewed as using the instruction “Ignore clothes" to retrieve and the traditional ReID can be seen as utilizing the instruction "Do not change clothes" to retrieve images of individuals wearing the same clothing. As another example, clothes template-based clothes-changing ReID can utilize a clothes template image as instruction. 
% For instance, clothes-changing ReID can be conceptualized as employing the instruction "Ignore clothes" to retrieve images of individuals whose clothing has been changed. Traditional ReID, on the other hand, can be seen as utilizing the instruction "Do not change clothes" to retrieve images of individuals wearing the same clothing. 
% Visible-infrared ReID can be perceived as using the instruction "Cross modality" to guide the model in retrieving images of individuals across different modalities. Clothes template-based clothes-changing ReID can utilize a clothing template image as instruction, with the model tasked to retrieve images of individuals wearing the template clothing. Language-instructed ReID involves the use of several sentences describing pedestrian attributes as instructions, with the model required to retrieve images of individuals described by the instructions. Text-to-image ReID simply employs the description sentence as input, without image data, to retrieve images.
The proposed instruct-ReID task offers three significant advantages: \ul{easy deployment, improved performance, and easy extension to new ReID tasks}. First, it enables cost-effective and convenient deployment in real-world applications. Unlike existing ReID approaches limited to specific tasks, instruct-ReID allows for utilizing a single model for various ReID scenarios, which is more practical in real applications. Second, as all existing ReID tasks can be considered special cases of instruct-ReID, we can unify the training sets of these tasks to exploit the benefits of more data and diverse annotations across various tasks -- leading to enhanced performance. Third, instruct-ReID introduces a new language-instructed ReID task which requires the model to retrieve persons following language instructions. 
This setting is practical for real-world applications, enabling customers to retrieve specific images through language descriptions. 
For example, customers can retrieve a woman wearing a black coat using one of her pictures and language instructions “Wear black coat".

\begin{figure*}
    \centering
    \includegraphics[width=\linewidth]{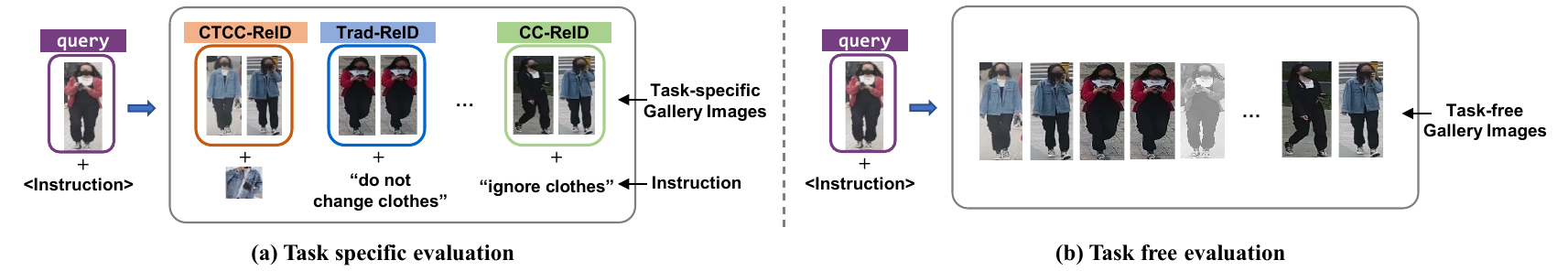}
    \caption{The comparison between task-specific and task-free evaluation settings. Task-free evaluation setting offers a more uniformly flexible evaluation approach.}
    \label{fig:evaluation_setting}
\end{figure*}

To facilitate research in Instruct-ReID task, we introduce a new benchmark called OmniReID, derived from 12 datasets \footnote{The discussion on ethical risks is provided in supplementary materials.} representing 6 distinct ReID tasks as described in our previous conference paper. In this paper, we extend OmniReID to OmniReID++, which has extra training data with 99,174 images, 5,221 identities, and 128,413 instructions. Compared to the OmniReID benchmark, the OmniReID++ has the following three improvements: 
\begin{itemize}
    \item The OmniReID++ benchmark emphasizes richer \emph{\textbf{diversity}} instruction annotations by incorporating images from a broader range of domains, which extends the scenarios from surveillance and synthetic games to include scenes from movies and internet videos. As shown in Tab.~\ref{tab:diversity}, OmniReID++ benchmark supplements instructions from winter environments and provides annotations for personnel across seasons and years. OmniReID++ expands the annotated scenes to include mountains, rivers, parks, and movie scenes. The diversity ensures that the trained models are robust and can effectively handle ReID tasks in various real-world scenarios.
    \item The OmniReID++ benchmark achieves improved \emph{\textbf{comprehensiveness}} by offering evaluation datasets to assess various ReID tasks. Specifically, in this paper, we present two evaluation settings along with the large-scale OmniReID++ benchmark, which facilitates evaluating the generalization ability of diverse ReID methods and can be outlined as follows: \\
    \underline{\emph{Task-specific evaluation setting.}} In existing benchmarks, the test datasets are designed for specific task scenarios, \emph{e.g.,} Market1501~\cite{zheng2015person} for Trad-ReID and PRCC~\cite{yang2019person} for CC-ReID, evaluating the performance of the model on the corresponding task. The OmniReID++ adopts this approach as a task-specific evaluation setting and utilizes 10 publicly available test sets covering 6 ReID tasks. As shown in \textcolor{black}{Fig.~\ref{fig:evaluation_setting} (a)}, both images in the query and gallery are associated with a task-specific instruction description, the model is evaluated separately on each task dataset and requires repeated testing with multiple datasets to evaluate the performance on multiple tasks. \\
    \underline{\emph{Task-free evaluation setting.}} \textcolor{black}{When evaluating the generalization ability of the model across multiple retrieval tasks jointly, task-specific instructions for each gallery image are discarded. Therefore, we propose a novel task-free evaluation setting as an extension of our previous conference paper. As shown in Fig. 2(b), the test set maintains a consistent gallery image input for all tasks, with each image in the gallery set being task-agnostic. When extracting gallery data features, only the image features are extracted for various ReID tasks. The gallery features can be fixed and extracted only once in applications, thereby avoiding re-extraction across various application scenarios in the task-specific evaluation setting, which saves computational resources.}
    \item The OmniReID++ benchmark proposes enhanced \emph{\textbf{exhaustiveness}} metric by introducing a novel \emph{\textbf{mAP$\tau$}}, which evaluates both the correctness of identity and the consistency with the instructions. The mAP$\tau$ provides a measurement to evaluate the alignment of retrieval results with the finer-grained retrieval intentions, which is different from the metrics in previous ReID tasks, \emph{e.g.,} rank-1 accuracy (R1) and mean average precision (mAP), that only measure identity correctness of retrieval images. Specifically, retrieval results with both the correct identity and its similarity with the given instruction exceeding the threshold $\tau$ will be considered as the correct retrieval, and the mean average precision is computed based on the refined rank list.
\end{itemize}

\begin{table}
\caption{Comparison of OmniReID and OmniReID++. Considering the aspects of instruction season, time span, and scene diversity, OmniReID++ offers a richer variety compared to OmniReID.}
% The Instruction style describes the seasonal characteristics of clothing templates and language descriptions. OmniReID++ provides instructions for different seasons throughout the year. The Time span indicates the time span of the same person across different images, with OmniReID++ including individuals spanning across seasons and years. The Scene reflects the diversity of image domains. Here, 'stone' represents environments primarily composed of sand and stone, while 'river' represents environments dominated by rivers, lakes, and beaches. In comparison, OmniReID++ offers a richer variety of scenes.
\centering
\renewcommand\arraystretch{1.2}
\label{tab:diversity}
\resizebox{\linewidth}{!}{
\begin{tabular}{llcc} 
\hline
Dataset Feature & Attributes & OmniReID & OmniReID++  \\ 
\hline
\multirow{4}{*}{Season} & Spring & \checkmark & \checkmark     \\
& Summer & \checkmark   & \checkmark     \\
& Autumn &  \checkmark  & \checkmark     \\
& Winter &    & \checkmark     \\ 
\hline
\multirow{4}{*}{Time Span}   & Cross day    & \checkmark  & \checkmark     \\
& Cross month  & \checkmark  & \checkmark     \\
& Cross season &    & \checkmark     \\
& Cross year   &    & \checkmark     \\ 
\hline
\multirow{8}{*}{Scene}  & Indoor & \checkmark  & \checkmark     \\   
& Street & \checkmark  & \checkmark     \\
& Mountain &   & \checkmark     \\
% & Stone &  & \checkmark     \\
& River &   & \checkmark     \\
& Park &   & \checkmark     \\
& Movie  &    & \checkmark     \\
& Synthetic    & \checkmark  & \checkmark     \\
\hline
\end{tabular}
}
\end{table}

To address the instruct-ReID task, we present an Instruct-ReID Model (IRM) in our previous conference version that can, for the first time, handle the major 6 ReID tasks listed in \textcolor{black}{Fig.~\ref{fig:overview} (a)} with a dual-branch framework. Unlike the typical triplet loss~\cite{schroff2015facenet} that only defines positive/negative pairs by identities, our IRM incorporates a novel adaptive triplet loss to learn a metric space that preserves identity and instruction similarities. Specifically, we design an adaptive margin between two query-instruction pairs based on instruction similarities to pull features with similar instructions close and push features with different instructions apart. Although the previous IRM has achieved significant success in unifying the multiple ReID tasks within one framework, it can not be applied in the newly proposed task-free evaluation setting for the following two reasons. First, the gallery features of various ReID tasks are extracted from different modules of IRM, \emph{e.g.,} Trad-ReID from the output of the attention module and T2I-ReID from the image encoder. However, in the task-free evaluation setting, all images in the gallery set are task-agnostic, making it impractical to extract retrieval features from different model parts based on the task type of gallery images. Second, IRM extracts gallery images and specific instruction features for retrieval comparison, while these task-specific instructions are unavailable in the task-free evaluation setting. To address the challenges, we introduce a new method called IRM++, which employs simple and unified gallery features for various ReID tasks. Concretely, unlike IRM, IRM++ obtains the gallery features by only extracting the image features from diverse tasks using an image encoder, without introducing task-specific multimodality instruction inputs. Furthermore, IRM++ introduces two memory banks in a contrastive learning manner to effectively supervise the model learning. Unlike triplet loss-based methods that include only one negative sample per triplet, memory bank contrastive learning offers multiple negative samples for each iteration, enhancing the discriminative representation learning of the data.

% ----- waiting experiment ----- %
We validate the effectiveness of IRM and IRM++ using ReID-specific pretraining~\cite{zhu2022pass} and OmniReID++ benchmark training data without finetuning, demonstrating consistent improvement over previous models across 10 datasets covering 6 ReID tasks. For task-specific evaluation setting, IRM improves \textbf{+7.7\%}, \textbf{+0.6\%}, \textbf{+0.5\%} mAP on CUHK03, MSMT17, Market1501 for traditional ReID, \textbf{+6.4\%}, \textbf{+11.2\%}, \textbf{+7.1\%} mAP on PRCC, LTCC, VC-Clothes for clothes-changing ReID when using RGB images only, \textbf{+11.7\%} mAP on COCAS+ real2 for clothes template based clothes changing ReID, \textbf{+4.3\%} mAP on LLCM for visible-infrared ReID, \textbf{+2.6\%} mAP on CUHK-PEDES for text-to-image ReID, \textbf{+24.9\%} mAP on COCAS+ real2 for the new language-instructed ReID.  For task-free evaluation setting, IRM++ achieves improvements of \textbf{+0.3\%}, \textbf{+0.7\%}, and \textbf{+7.0\%} mAP on Market1501, MSMT17, and CUHK03 for traditional ReID tasks, \textbf{+1.8\%}, \textbf{+6.6\%}, and \textbf{+5.9\%} mAP on PRCC, VC-Clothes, and LTCC for clothes-changing ReID. Additionally, for clothes template-based clothes-changing ReID and our newly introduced language-instructed ReID task, we observe a \textbf{+5.3\%} mAP and a \textbf{+22.4\%} mAP enhancement on COCAS+ real2 dataset. Furthermore, our model achieves \textbf{+4.2\%} mAP on LLCM for visible-infrared ReID and \textbf{+2.5\%} mAP on CUHK-PEDES for text-to-image ReID.

In summary, the contributions of this paper are three folds. (1) We propose a new instruct-ReID task, where existing traditional ReID, clothes-changing ReID, clothes template-based clothes-changing ReID, visible-infrared ReID, text-to-image ReID, and language-instructed ReID can be viewed as special cases. (2) To facilitate research on instruct-ReID, we establish a large-scale and comprehensive benchmark: OmniReID++, covering the existing 6 ReID tasks. (3) We propose an adaptive triplet loss in IRM to supervise the feature distance of two query-instruction pairs to consider identity and instruction alignments. We utilize memory bank assisted learning in IRM++ to introduce powerful supervision in model training, obtaining better performance for the task-free evaluation setting. 

\textbf{Difference from the conference paper.}
A preliminary version of this paper is presented in the Instruct-ReID Model (IRM)~\cite{he2023instructreid}. This paper extends the previous study with four major improvements. 
1) We extend OmniReID to OmniReID++, leveraging more diverse data sources. This expansion places a wider variety of domains, ranging from surveillance and synthetic games to movies and internet videos. This broader dataset ensures that the models trained on OmniReID++ are more adaptable to real-world scenarios.
% \textcolor{black}{2) We propose a more comprehensive evaluation method for the instruct-ReID task with a novel task-free evaluation setting. The novel evaluation setting unifies the gallery image features of various ReID tasks, which is more aligned with the requirements of retrieving from gallery sets without task specifications in real-world applications.}
2) We propose a new task-free evaluation setting that is more aligned with real-world inference scenarios. The novel evaluation setting unifies the gallery retrieval features of various ReID tasks and avoids introducing task-specific instructions for gallery images, which addresses the requirements of retrieving from task-agnostic gallery images and further enhances the comprehensiveness of the evaluation methodology.
3) We introduce a novel evaluation metric, mAP$\tau$, to provide a more precise assessment of whether retrieval results meet the retrieval target. The novel mAP$\tau$ measures both the correctness of identity and the consistency of retrieval results with the instructions in the rank list, thereby integrating a finer discrimination method beyond identity criteria.
4) We propose IRM++ with memory bank assisted learning for instruct-ReID task and conduct additional experiments to evaluate the effectiveness of the IRM++. Experiments on the task-free evaluation setting show that IRM++ achieves significant improvements based on OmniReID++ benchmark, \emph{e.g.,} \textbf{+9.0\%} and \textbf{+5.8\%} mAP on CTCC-ReID and LI-ReID tasks compared to IRM.

\section{Related Work}
\subsection{Person Re-identification}
Person re-identification aims to retrieve the same images of the same identity with the given query from the gallery set. To support the ReID task on all-weather application, various tasks are conducted on the scenarios with changing environments, perspectives, and poses~\cite{ye2021deep,chen2017person,zheng2016person,chen2018improving,zheng2020dual}. Traditional ReID mainly focuses on dealing with indoor/outdoor problems when the target person wears the same clothes. Recently, to extend the application scenarios, clothes-changing ReID (CC-ReID)~\cite{yu2020cocas} and clothes template based clothes-changing ReID (CTCC-ReID)~\cite{li2022cocas+} are proposed. While CC-ReID forces the model to learn clothes-invariant features~\cite{huang2021clothing,gu2022clothes}, CCTC-ReID further extracts clothes template features~\cite{li2022cocas+} to retrieve the image of the person wearing template clothes. To capture person's information under low-light environments, visible-infrared person ReID (VI-ReID) methods~\cite{zhang2022fmcnet,yang2022augmented,yang2022learning} retrieve the visible (infrared) images according to the corresponding infrared (visible) images. In the absence of the query image, GNA-RNN~\cite{li2017person} introduced the text-to-image ReID (T2I-ReID) task, which aims at retrieving the person from the textual description. However, existing tasks and methods focus on a single scenario, making it difficult to address the demands of cross-scenario tasks. In this paper, we introduce a new Instruct-ReID task, which can be viewed as a superset of the existing ReID tasks by incorporating instruction information into identification.
% \subsection{Instruction Tuning}
% Instruction Tuning was first proposed in natural language processing (NLP) to enable large language models (LLM) to execute specific tasks by following natural language instructions. LLM instruction-tuned models, \emph{e.g.}, FLAN-T5~\cite{chung2022scaling}, InstuctGPT~\cite{ouyang2022training}/ChatGPT~\cite{openai_chatgpt}, can effectively prompt the ability on zero- and few-shot transfer tasks. A few works have borrowed the idea from language to vision. Flamingo~\cite{alayrac2022flamingo}, BLIP-2~\cite{li2023blip}, and KOSMOS-1~\cite{huang2023language} learning with image-text pairs also show promising generalization on visual understanding tasks. While these LLM-based methods aim to generate convincing language responses following the image or language instructions, we focus on retrieving the correct person following the given instructions by tuning a vision transformer.
\subsection{Loss Function in ReID}
Mulitple loss function have been extensively studied during the development of the ReID research~\cite{ye2021deep}. Depending on the structure of the method design, there are three typical types of loss function: \textbf{(1) Identity loss} proposed in~\cite{zheng2017person, sun2017svdnet,zheng2017unlabeled, huang2018adversarially, zheng2019re,li2019unsupervised} treats each identity as a distinct class, thus turning the training phase of ReID as a classification process. Identity loss using cross-entropy loss to compute the difference between the labels and the predicted probability. It's convenient for training but easy to overfit when the labels are over-confident annotated~\cite{muller2019does}. 
\textbf{(2) Contrastive loss}~\cite{varior2016siamese, deng2018image,chen2022learning,rao2021self,su2020adapting,he2025adept} has been widely used in diverse ReID scenarios. It utilizes Euclidean distance to compute the pairwise distance of two input sample features. Contrastive loss encourages features of the same identities to come closer together while pushing features from different identities farther apart. 
% Similar to the contrastive loss, binary verification loss~\cite{zheng2017discriminatively, li2014deepreid} also deals with a pair of samples. It uses cross-entropy loss to classify two input samples into a positive or negative pair due to the distance between their features. 
The Contrastive loss can alleviate the impact of wrong classifications and noise during the clustering stage. Although contrastive loss can be sensitive to the choice of positive and negative samples during training, it faces the challenge of lacking a reference distance between samples in ReID task applications. 
\textbf{(3) Triplet loss}~\cite{hermans2017defense, wang2018resource, shi2016embedding,hou2021feature} is also an extensively explored  feature clustering method. It uses a retrieval ranking perspective to address the training process of the ReID model~\cite{ye2021deep}. Triplet loss takes one anchor sample, one positive sample, and one negative sample as input; based on the assumption that the distance between the anchor and the positive sample should be smaller than the distance between the anchor and the negative sample by a predefined margin, triplet loss optimizes the distance within the triplet samples during the training phase. 

Different loss functions are often combined to achieve better performance. For instance, a common approach involves linearly combining the identity loss with the triplet loss~\cite{sun2019learning}. In our paper, we employ identity loss, contrastive loss, and triplet loss to provide better optimization for instruct-ReID task. While triplet loss has achieved success in traditional ReID tasks, it fails to distinguish the two positive samples with distinct instructions. Different from previous methods using the vanilla triplet loss with a fixed margin, we proposed an adaptive triplet loss with a variable margin that takes the similarity of sample instructions into consideration. Our approach offers a new solution to the instruction-based retrieval task.

\subsection {Multimodal Retrieval}
Multimodel retrieval is a widely used approach that aligns information from multiple modalities to broaden the scope of model applications. In multimodel retrieval, unimodal encoders always encode different modalities for retrieval tasks. For instance, CLIP~\cite{radford2021learning}, VideoCLIP~\cite{xu2021videoclip}, COOT~\cite{ging2020coot} and MMV~\cite{alayrac2020self} utilize contrastive learning to align the multimodal features for pre-training. Other techniques like HERO~\cite{li2020hero}, Clipbert~\cite{lei2021less}, Vlm~\cite{xu2021vlm}, bridgeformer~\cite{ge2022bridgeformer} and UniVL~\cite{luo2020univl} focus on merging different modalities for retrieval tasks to learn a generic representation. 
Although there have been numerous studies on multimodal retrieval, most are concentrated on language-vision pretraining or video retrieval, leaving the potential of multimodal retrieval for person ReID largely unexplored. This paper aims to investigate this underexplored area to retrieve anyone with information extracted from multiple modalities.

\subsection{Memory Bank}
Memory bank is a non-parametric buffer that stores the key features during training. 
Memory bank and memory-based learning have been widely explored in supervised learning~\cite{wang2020cross}, semi-supervised and unsupervised learning~\cite{wu2018unsupervised,tarvainen2017mean}. ~\cite{wu2018unsupervised} introduced a memory bank in instance discrimination pretext task, storing the representations of all samples in the dataset, providing sufficient negative samples to the feature clustering.  It has also been used in various tasks, such as few-shot video classification~\cite{zhu2020label}, semantic segmentation~\cite{fan2022memory}, video deblurring~\cite{ji2022multi} and so on. 
MAUM~\cite{liu2022learning} adopts memory bank in cross-modality person re-identification by leveraging model drift to provide hard positive references, thereby improving robustness against modality discrepancy and imbalance. 
In IRM++, we introduce a dual-branch memory bank to store the features for both instructions and images for each identity. Different from former MAUM using memory bank for better cross-modality association, IRM++ utilizes memory bank to enhance the model's capacity for instruction following.
% will have more

\section{OmniReID++ Benchmark}
To facilitate research on the Instruct-ReID task, we propose the OmniReID++ benchmark including a large-scale pretraining dataset based on 13 publicly available datasets with visual and language annotations. The comparison with the existing ReID benchmark is illustrated in Tab.~\ref{OmniReID benchmark}. The OmniReID++ has a larger dataset compared to existing benchmarks and the previous version of OmniReID benchmark. While MALS~\cite{yang2023towards} has more identities, its data is synthetic, and our OmniReID++ dataset shows greater diversity. 

\begin{table*}
    \renewcommand\arraystretch{1.1}
    \caption{Comparison of training subsets of different ReID datasets.}
    \label{OmniReID benchmark}
    \centering
    \footnotesize
    \resizebox{\linewidth}{!}{
    \begin{tabular}{l|cccccccc} 
    \toprule
    dataset    & image~ & ID & \textcolor{black}{scene} & \textcolor{black}{labeled} & \textcolor{black}{camera view} & \textcolor{black}{crop size} & \textcolor{black}{resolution} & domain  \\ 
    \midrule
    MSMT17~\cite{wei2018person}     & 30,248  & 1,041 & \textcolor{black}{15} & \textcolor{black}{yes} & \textcolor{black}{fixed} & \textcolor{black}{vary} & \textcolor{black}{fixed} & indoor/outdoor      \\
    % Duke~\cite{zheng2017unlabeled} & 16522  & 702  & outdoor \\
    %CUHK03~\cite{li2014deepreid} & 7,365  & 767  & indoor \\
    Market1501~\cite{zheng2015person} & 12,936  & 751 & \textcolor{black}{6} & \textcolor{black}{yes} & \textcolor{black}{fixed} & \textcolor{black}{128$\times$64} & \textcolor{black}{fixed} & outdoor \\
    PRCC~\cite{yang2019person} & 17,896  & 150 & \textcolor{black}{3} & \textcolor{black}{yes} & \textcolor{black}{fixed} & \textcolor{black}{vary} & \textcolor{black}{fixed} & indoor  \\
    %VC-Clothes~\cite{wan2020person} & 9,449   & 256  & synthetic     \\
    COCAS+ Real1~\cite{li2022cocas+}      & 34,469  & 2,800 & \textcolor{black}{30} & \textcolor{black}{yes} & \textcolor{black}{fixed} & \textcolor{black}{256$\times$128} & \textcolor{black}{fixed} & indoor/outdoor      \\ 
    LLCM~\cite{zhang2023diverse}      & 30,921  & 713 & \textcolor{black}{9} & \textcolor{black}{yes} & \textcolor{black}{fixed} & \textcolor{black}{vary} & \textcolor{black}{fixed} & indoor/outdoor      \\ 
    %CUHK-PEDES~\cite{li2017person}      & 34,054  & 11,003 & indoor/outdoor      \\ 
    LaST~\cite{shu2021large}      & 71,248  & 5,000 & \textcolor{black}{-} & \textcolor{black}{yes} & \textcolor{black}{dynamic} & \textcolor{black}{vary} & \textcolor{black}{dynamic} & indoor/outdoor \\
    MALS~\cite{yang2023towards}      & 1,510,330  & 1,510,330 & \textcolor{black}{-} & \textcolor{black}{yes} & \textcolor{black}{dynamic} & \textcolor{black}{531$\times$208} & \textcolor{black}{fixed} & synthesis \\
    LUPerson-T~\cite{shao2023unified}      & 957,606  & - & \textcolor{black}{-} & \textcolor{black}{pseudo} & \textcolor{black}{dynamic} & \textcolor{black}{vary} & \textcolor{black}{dynamic}  & indoor/outdoor \\
    \hline\hline
    % OmniReID   & 4,973,044 & 328,604 & indoor/outdoor/synthetic  \\
    OmniReID++   & 5,072,218 & 333,825 & \textcolor{black}{$\textgreater$21,697} & \textcolor{black}{yes} & \textcolor{black}{dynamic} & \textcolor{black}{vary} & \textcolor{black}{dynamic} & indoor/outdoor/synthetic  \\
    % 4973044 = 30248+7365+12936+34469+17896+9449+9423+30921+28566+4791771
    % 5072218 = 30248+7365+12936+34469+17896+9449+9423+30921+28566+4791771+71248+5336+22590
    % 328604  = 5842+713+312321+9728 
    % 333825  = 5842+713+312321+9728+5000+40+181
    \bottomrule
    \end{tabular}
    }
\end{table*}

%-------------------------------------------------------------------------
\noindent \textbf{Data Collection.}
To enable all-purpose person ReID, we collect massive public datasets from various domains and use their training subset as our training subset, including Market1501~\cite{zheng2015person}, MSMT17~\cite{wei2018person}, CUHK03~\cite{li2014deepreid} from traditional ReID, PRCC~\cite{yang2019person}, VC-Clothes~\cite{wan2020person}, LTCC~\cite{qian2020long}, LaST~\cite{shu2021large}, NKUP~\cite{wang2020benchmark} and NKUP+~\cite{liu2022long} from clothes-changing ReID, LLCM~\cite{zhang2023diverse} from visible-infrared ReID, CUHK-PEDES~\cite{li2017person}, SYNTH-PEDES~\cite{zuo2023plip} from text-to-image ReID, COCAS+ Real1~\cite{zhang2023diverse} from clothes template based clothes-changing ReID and language-instructed ReID, forming 5,072,218 images and 333,825 identities. To fairly compare our method with state-of-the-art methods, the trained models are evaluated on LTCC, PRCC, VC-Clthoes, COCAS+ Real2, LLCM, CUHK-PEDES, Market1501, MSMT17, and CUHK03 test subsets without fine-tuning. We present all dataset statistics of OmniReID++ in the supplementary materials. 

\begin{figure}
  \centering
  \includegraphics[width=\linewidth]
  {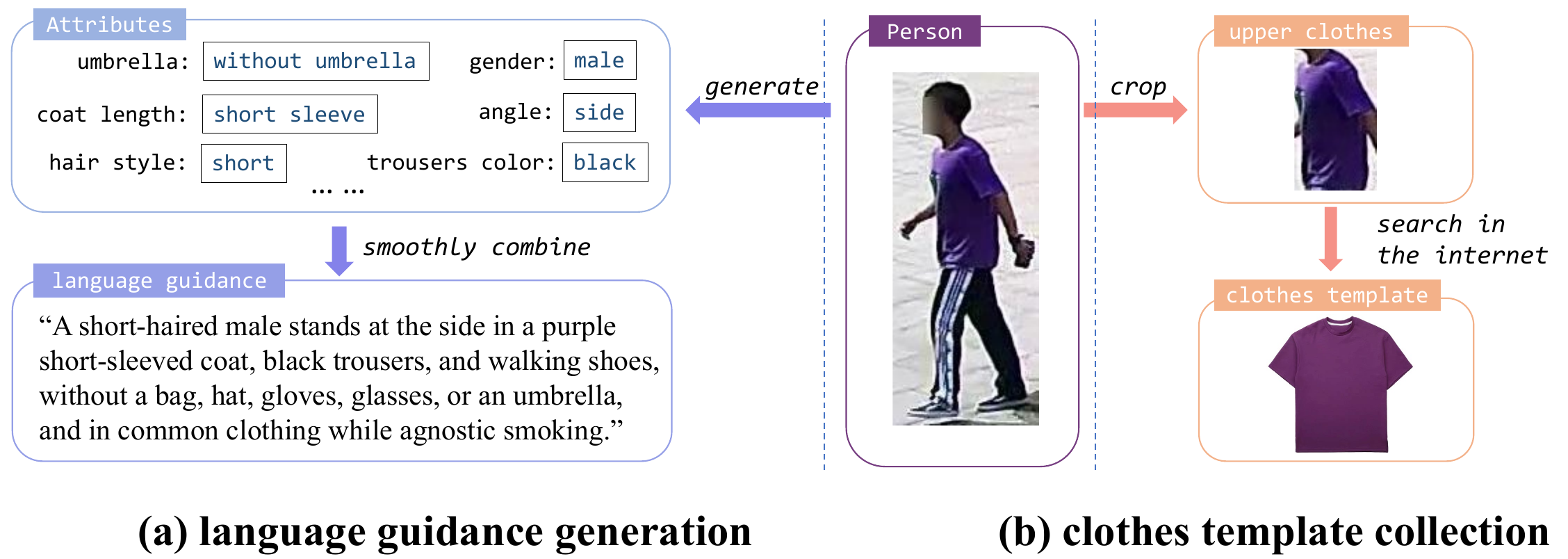}
  \caption{\textbf{(a)} We generate attributes for a person and then transform attributes into sentences by a large language model. \textbf{(b)} We crop upper clothes and search them online for clothes templates.}
  \label{fig:attr}
\end{figure}

\noindent \textbf{Language Annotation Generation.}
To generate language instruction for language-instructed ReID, we annotate COCAS+ Real1, LTCC, PRCC, LaST, NKUP, NKUP+ and COCAS+ Real2 with language description labels. Similar to Text-to-Image ReID datasets, \emph{e.g.,} CUHK-PEDES~\cite{li2017person}, language annotations in OmniReID++ are several sentences that describe the visual appearance of pedestrians. We divide our annotation process into \emph{pedestrian attribute generation} and \emph{attribution-to-language transformation}. \\
\underline{\emph{Pedestrian Attribute Generation.}} To obtain a comprehensive and varied description of an individual, we employ an extensive collection of attribute words describing a wide range of human visual characteristics. The collection contains 20 attributes and 92 specific representation words, including full-body clothing, hair color, hairstyle, gender, posture, and accessories such as umbrellas or satchels. Professional annotators manually label all the pedestrian attributes. We provide a practical illustration with the attribute collection in \textcolor{black}{Fig.~\ref{fig:attr} (a)}. By utilizing instructions on the well-defined attribute combination, models can further enhance ability to identify the target person. \\
\underline{\emph{Attribute-to-Language Transformation.}}
Compared with discrete attribute words, language is more natural for consumers. To this end, we transform these attributes into multiple sentences using the Alpaca-LoRA~\cite{zhang2023llama} large language model. Specifically, we ask the Alpaca-LoRA with the following sentences: “Generate sentences to describe a person. The above sentences should contain all the attribute information I gave you in the following." 
The generated annotations are carefully checked and corrected manually to ensure the correctness of the language instructions. All detailed pedestrian attributes and more language annotations are presented in the supplementary materials.

\noindent \textbf{Visual Annotation Generation.} Visual annotations are images that describe the characteristics of pedestrians. In this paper, we select clothes as the visual annotations because they are viewed as the most significant visual characteristics of pedestrians. To get high-quality visual annotations, we first crop the upper clothes from the source images (Clothes Copping) and search on the internet to get the most corresponding clothes-template images (Clothes-template Crawling) as visual annotations. Since each person wears the same clothes in traditional ReID datasets, we annotate the clothes-changing PRCC, LTCC and LaST datasets where each person wears multiple clothes to ease the burden of annotations. \textcolor{black}{Fig.~\ref{fig:attr} (b)} shows the detailed process. \\
\underline{\emph{Clothes Cropping.}} We use a human parsing model SCHP~\cite{li2020self} to generate the segmentation mask of the upper clothes and then crop the corresponding rectangle upper clothes patches from the original images. These bounding boxes of upper clothes are then manually validated. \\
\underline{\emph{Clothes-template Crawling.}} Given all cropped upper clothes from images in OmniReID datasets, we crawl the templates of these clothes from shopping websites\footnote{https://world.taobao.com/, https://www.17qcc.com/}. The top 40 matching clothes templates are downloaded when we crawl each cropped upper clothes. The one with the highest image quality is manually selected.

\begin{figure}
    \centering
    \includegraphics[width=\linewidth]{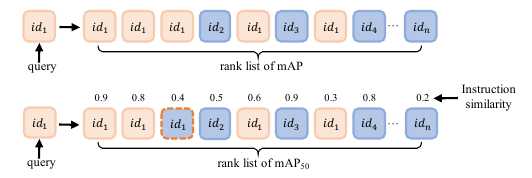}
    \caption{Illustration of the difference between metric mAP and mAP$\tau$. Light orange in the rank list indicates positive and negative samples in light blue. In the example of mAP$\tau$, we set the threshold to 0.5 and mark with an orange dashed box where the identity matches the query but the instruction similarity does not reach the threshold and is corrected to a negative sample.}
    \label{fig:metric}
\end{figure}

\noindent \textbf{Training Setting} \label{sec:traning_datasets}
% Describe the categories of datasets and why we choose these datasets for training, briefly introduce typeciall datasets and rest in sup.
To enable all-purpose person ReID, we perform two training scenarios based on this built benchmark: 
(1) Single-task Learning (STL): Every task is trained and tested individually using the corresponding dataset.
(2) Multi-task Learning (MTL): To acquire one unified model for all tasks, the model is optimized by joint training of the six ReID tasks with all the training datasets. The trained network is then evaluated for different tasks on various datasets.

\noindent \textbf{Evaluation Setting}
% \label{sec:test_datasets}
To validate the effectiveness of ReID methods, we propose two testing methodologies along with the OmniReID++ benchmark: 
(1) Task-specific evaluation: We independently utilize each test dataset included in the OmniReID++ benchmark to validate the performance of a ReID model on that specific task and dataset. When extracting gallery features for person retrieval, each image in the gallery set is associated with a task-specific instruction. 
(2) Task-free evaluation: In this setting, we consolidate all test datasets from the OmniReID++ benchmark into a single test dataset, which includes query images and instructions representing different task retrieval targets. During inference, the gallery datasets from various sub-datasets are original images without distinction. This approach better simulates the inference scenario of the model in real-world applications.

\noindent \textbf{Evaluation Metric}
OmniReID++ employs the commonly used CMC (Cumulative Matching Characteristic) metric along with mAP (mean Average Precision) as the traditional evaluation metrics. Besides, we also introduce a novel mAP$\tau$ metric that provides a method to evaluate both the accuracy of person retrieval identity and the compliance with instruction target for instruct-ReID task. The calculation of mAP$\tau$ is defined as
\begin{equation} \small
    \begin{split}
    \mathbf{mAP\tau}=\frac{1}{Q} \sum_{q=1}^{Q} \sum_{k=1}^{n} \frac{\Psi^{\tau}(l_q,l^{q}_k)}{kn_q} \sum_{i=1}^{k} \Psi^{\tau}(l_q,l^{q}_i)
    % \mathbf{Prob}(s^{a}_{i}) /  d(\mathbf{F}^{a}_{i},\mathbf{F}^{r_{1}}_{i}) + \\ \left ( \beta_1 - \beta_2  \right )m - d(\mathbf{F}^{a}_{i},\mathbf{F}^{r_{2}}_{i})  ]  \}_{+} 
    % \mathcal{F}_l(\mathbf{F}_{l-1}, \mathbf{F}_{\mathbf{T}})
    \end{split}
\end{equation}
where $Q$ is the number of query images, $n$ is the maximum computation length set in the $q$-th image retrieval rank list, $n_q$ is the correct returned results number, $\Psi^{\tau}(\cdot)$ denotes the matching evaluator and the calculation process for the $q$-th query $l_q$ with a $k$-th retrieval result $l^{q}_k$ can be defined as
\begin{equation} \small
    \begin{split}
    \Psi^{\tau}(l_q,l^{q}_k) = \mathbb{I}\left(y_q=y_k\right) \cdot \mathbb{I}\left(\mathbf{Cos} \left \langle I_q, I^{q}_k \right \rangle \geq \tau\right)
    \end{split}
\end{equation}
where $\mathbb{I}(\cdot)$ is the indicator function, $y_q$ or $y_k$ is the identity label, $I_q$ or $I^{q}_k$ is instruction and $\mathbf{Cos\left \langle \cdot \right \rangle}$ is the cosine similarity function. The matching evaluator is 1 if the $k$-th or $i$-th prediction is correct for the $q$-th query $l_q$ after applying the threshold $\tau$ and 0 for otherwise.
The illustration and distinction between mAP and mAP$\tau$ are described in Fig.~\ref{fig:metric}. In the calculation process of mAP, images in the retrieval rank list are categorized as positive or negative depending on whether their identities match that of the query image. However, in the proposed instruct-ReID task, we search the target images with both query images and the instructions. The retrieved images with the same identity may not necessarily match the instruction target. Hence, the new proposed mAP$\tau$ computes the similarity between each instruction of the gallery image and the query image, and samples in the rank list with instruction similarity below a predefined threshold $\tau$ are considered negative. We take the third sample (red dash boundary sample in Fig.~\ref{fig:metric}) in the mAP$_{50}$ rank list for example, where mAP$_{50}$ denotes a threshold $\tau$ set at 0.5. Despite sharing the same identity as the query image, its instruction similarity is below 0.5, leading to its classification as a negative sample for calculation purposes. Compared to the traditional mAP metric, the mAP$\tau$ metric provides a more precise reflection of whether retrieval results align with the instruction description, even in individual text-to-image tasks.

\begin{figure*}
  \centering
  \includegraphics[width=0.95\linewidth]{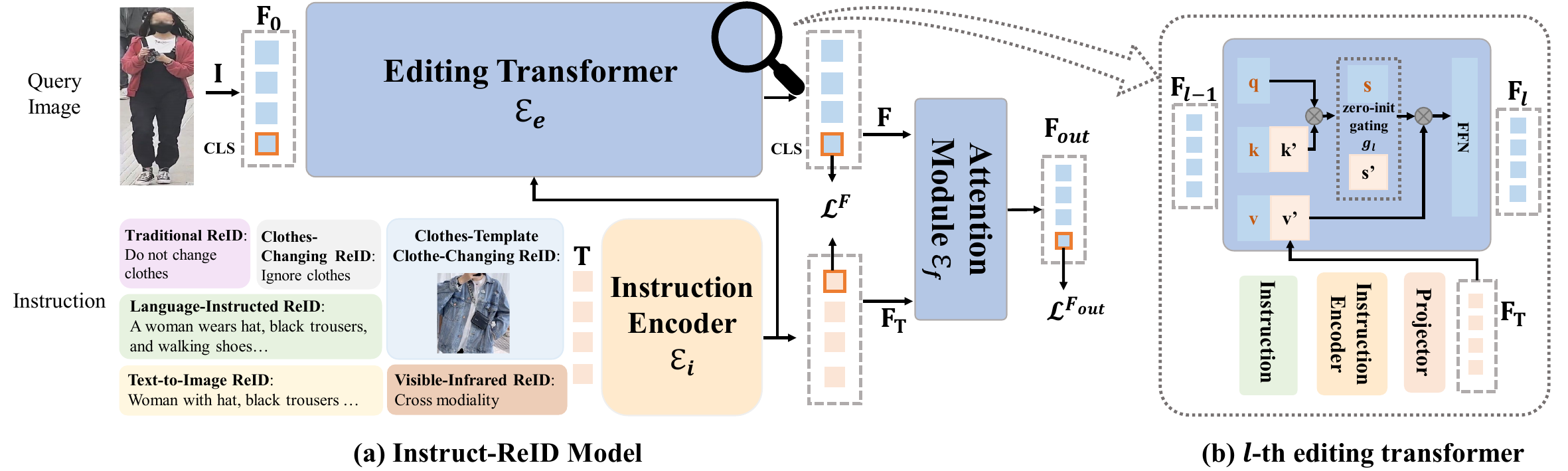}
  \caption{The overall architecture of the proposed method. The instruction is fed into the instruction encoder to extract instruction features (a). The features are then propagated into the editing transformer (b) to capture instruction-edited features. We exploit adaptive triplet loss and identification loss to train the network. For the testing stage, we use the CLS token for retrieval. }
  \label{fig:method}
\end{figure*}
% \vspace{-0.2em}

\section{Instruct ReID Methodology}
% model architechure; pipeline
In this section, we present our proposed instruct-ReID model, which includes IRM for the task-specific evaluation setting and IRM++ for the task-free evaluation setting. We begin by introducing the instruction generation (Sec.~\ref{sec:instruct_generation}) for different ReID tasks which provides specific instructions for query retrieval guidance for both IRM and IRM++. Then the architecture of the proposed Instruct ReID model IRM is discussed in (Sec.~\ref{sec:IRM}). The overview of IRM is shown in Fig.~\ref{fig:method}, which consists of three parts: an editing transformer $\mathcal{E}_{e}$ (Sec.~\ref{sec:edit_transformer}) and an instruction encoder $\mathcal{E}_{i}$ which is a visual language model, and an attention module $\mathcal{E}_{f}$. Instead of traditional triplet loss, IRM adopts a newly proposed adaptive triplet loss $\mathcal{L}_{atri}$ (Sec.~\ref{sec:adaptive_triplet_loss}) to supervise the model training and the overall loss function is introduced for all ReID tasks in (Sec.~\ref{sec:overall_loss}).
Finally, we describe IRM++ in (Sec.~\ref{sec:IRM++}) which also employs an image encoder $\mathcal{E}_{e}$, an instruction encoder $\mathcal{E}_{i}$ and an attention module $\mathcal{E}_{f}$. IRM++ only extracts images features by the image encoder $\mathcal{E}_{e}$ as the general gallery features retrieval for various ReID tasks. Besides, we present a memory bank contrastive learning method to assist IRM++ learning, achieving better performance in the task-free evaluation setting than the triplet loss based training methods.
% \textcolor{red}{Finally, Based on IRM, IRM++ (Sec.~\ref{sec:IRM++}) step further towards efficient retrieval and obtain better performance on task-free evaluation setting.}

\subsection{Instruction Generation} \label{sec:instruct_generation}

In our proposed instruct-ReID task, the model must retrieve the images that describe the same person following the instructions. By designing different instructions, our instruct-ReID can be specialized to existing ReID tasks, \emph{i.e.,} traditional ReID, clothes-changing ReID, clothes template based clothes-changing ReID, visual-infrared ReID, text-to-image ReID, and language-instructed ReID. We show the current instructions and leave the exploration of better instructions for instruct-ReID to future research.

\noindent \textbf{Traditional ReID:} Following Instruct-BLIP~\cite{instructblip},  we generate \underline{20} instructions\footnote{See all available instruction sentences in supplementary materials.} from GPT-4 and randomly select one, \emph{e.g.,} “Do not change clothes.", during training. The model is expected to retrieve images of the same person without altering image attributes, such as clothing. 
\begin{mdframed}[backgroundcolor=gray!10]
\textbf{\#\#\# Query image}:\{Query image\} \\ \quad \textbf{\#\#\# Instruction}: “Do not change clothes."$^3$ \\
\textbf{\#\#\# Target image}:\{Output\}
\end{mdframed}
\textbf{Clothes-changing ReID:} Simiar to Traditional ReID, the instruction is the sentence chosen from a collection of \underline{20} GPT-4 generated sentences$^3$, \emph{e.g.,} “With clothes changed". The model should retrieve images of the same person even when wearing different outfits. 
\begin{mdframed}[backgroundcolor=gray!10]
\textbf{\#\#\# Query image}:\{Query image\} \\ \quad \textbf{\#\#\# Instruction}: “With clothes changed."$^3$ \\
\textbf{\#\#\# Target image}:\{Output\}
\end{mdframed}
\textbf{Clothes template based clothes-changing ReID:} The instruction is a clothes template for a query image while a cropped clothes image for a target image. The model should retrieve images of the same person wearing the provided clothes. We provide more examples for training and test in the supplementary materials.
\begin{mdframed}[backgroundcolor=gray!10]
\textbf{\#\#\# Query image}:\{Query image\} \\ \quad \textbf{\#\#\# Instruction}:\{Any clothes template image\}\\
\textbf{\#\#\# Target image}:\{Output\}
\end{mdframed}
\textbf{Visual-Infrared ReID:} The instruction is the sentence chosen from a collection of \underline{20} GPT-4 generated sentences$^3$, \emph{e.g.,} “Retrieve cross modality image". The model should retrieve visible (infrared) images of the same person according to the corresponding infrared (visible) images. 
\begin{mdframed}[backgroundcolor=gray!10]
\textbf{\#\#\# Query image}:\{Query image\} \\ \quad \textbf{\#\#\# Instruction}: “Retrieve cross modality image."$^3$ \\
\textbf{\#\#\# Target image}:\{Output\}
\end{mdframed}
\textbf{Text-to-Image ReID:} The instruction is the describing sentences, and both images and text are fed into IRM during the training process. While in the inference stage, the image features and instruction features are extracted separately. \footnote{Image and instruction features are extracted separately in test stage.} 
\begin{mdframed}[backgroundcolor=gray!10]
\textbf{\#\#\# Image}:\{Image\}$^4$ \\ \quad \textbf{\#\#\# Instruction}:\{Sentences describing pedestrians\}$^4$\\
\textbf{\#\#\# Target image}:\{Output\}
\end{mdframed}
\textbf{Language-instructed ReID:} The instruction is several sentences describing pedestrian attributes. We randomly select the description languages from the person images in gallery and provide to query images as instruction. The model is required to retrieve images of the same person described in the provided sentences. We provide more examples for training and test in the supplementary materials.
\begin{mdframed}[backgroundcolor=gray!10]
\textbf{\#\#\# Query image}:\{Query image\} \\ \quad \textbf{\#\#\# Instruction}:\{Sentences describing pedestrians\}\\
\textbf{\#\#\# Target image}:\{Output\}
\end{mdframed}

\subsection{Model Architecture of IRM} \label{sec:IRM}
As illustrated in Fig.~\ref{fig:method} (a), the pipeline of IRM can be summarized as follows: given an instruction $\mathbf{T}$ associated with a query image $\mathbf{I}$, IRM obtains instruction features $\mathbf{F}_\mathbf{T}$ using the instruction encoder $\mathcal{E}i$ which is a visual language model. These extracted instruction features $\mathbf{F}_\mathbf{T}$, along with query image tokens, are fed into the designed editing transformer $\mathcal{E}_e$ to obtain features $\mathbf{F}$ edited by instructions. Moreover, both the instruction features $\mathbf{F}_\mathbf{T}$ and $\mathbf{F}$ are fed into an attention module $\mathcal{E}_f$ to efficiently integrate features of the query image and instruction using the attention mechanism within the stacked transformer layers. Finally, the output embeddings $\mathbf{F}_{out}$ of the attention module are utilized for person image retrieval.

\subsubsection{Editing Transformer} \label{sec:edit_transformer} 

The editing transformer consists of $L$ zero-init transformer layers $\mathcal{E}_e = \{\mathcal{F}_1, \mathcal{F}_2, ..., \mathcal{F}_L\}$, which can leverage the instruction to edit the feature of query images. Given the $l$-th zero-init transformer, the output feature $\mathbf{F}_l$ can be formulated as 
\begin{equation} \small
    \mathbf{F}_l = \mathcal{F}_l(\mathbf{F}_{l-1}, \mathbf{F}_{\mathbf{T}}),
\end{equation}
where $\mathbf{F}_\mathbf{T} = \mathcal{E}_i(\mathbf{T})$ is the instruction feature extracted by the instruction encoder $\mathcal{E}_i$ and $\mathbf{F}_{l-1}$ is the output feature of $(l\!-\!1)$-th zero-init transformer layer. Here, we define the initial input $\mathcal{F}_0$ of the first layer as $\mathbf{F}_0 = [\mathbf{f}_0^{\text{CLS}}, \mathbf{f}_0^{1}, \mathbf{f}_0^{2}, ..., \mathbf{f}_0^{N}],$
% \begin{equation}
%     \mathbf{F}_0 = [\mathbf{f}_0^{\text{CLS}}, \mathbf{f}_0^{1}, \mathbf{f}_0^{2}, ..., \mathbf{f}_0^{N}], 
% \end{equation}
where $\mathbf{f}^{\text{CLS}}_0$ is the appended [CLS] token,  $(\mathbf{f}_0^{1}, \mathbf{f}_0^{2}, ..., \mathbf{f}_0^{N})$ are the patch tokens of the query image and $N$ is the number of patches of the query image.

We show the detailed structure of each layer in the editing transformer in Fig.~\ref{fig:method} (b). Given the features $\mathbf{F}_{l-1}$ and instruction features $\mathbf{F}_{\mathbf{T}}$, the attention map $\mathbf{M}_l$ is defined as 
% \vspace{-1em}
\begin{equation} \small
    \mathbf{M}_l = \left[\text{Softmax}(\mathbf{S}_l), g_l \times \text{Softmax}(\mathbf{S'}_l)\right],
\end{equation}
where $g_l$ is the gating parameters initialized by zero. Here, $\mathbf{S}_l$ is  the attention map between queries and keys of input features and $\mathbf{S}'_l$ is the attention map between queries of input features and keys of instruction features. Mathematically, 
\begin{equation} \small
    \mathbf{S}_l = \mathbf{Q}_l\mathbf{K}_l^\top/\sqrt{C}, \mathbf{S}_l' = \mathbf{Q}_l\mathbf{K'}_l^\top/\sqrt{C},
\end{equation}
where a linear projection derives queries and keys, \emph{i.e.,} $\mathbf{Q}_l\!=\!\text{Linear}_q(\mathbf{F}_{l-1})$, $\mathbf{K}_l\!=\!\text{Linear}_k(\mathbf{F}_{l-1})$ and $\mathbf{K}'_l\!=\!\text{Linear}_{k'}(\mathbf{F}_{\mathbf{T}})$, respectively. $C$ is the feature dimension of query features. Finally, we calculate the output of the $l$-th layer by 
\begin{equation} \small
    \mathbf{F}_l = \text{Linear}_o(\mathbf{M}_l\left[\mathbf{V}_l, \mathbf{V}'_l\right]),
\end{equation}
where $\text{Linear}_o$ is the feed-forward network after the attention layer in each transformer block, $\mathbf{V}_l$ and $\mathbf{V}'_l$ are the values calculated by $\mathbf{V}_l=\text{Linear}_v(\mathbf{F}_{l-1})$ and $\mathbf{V}'_l=\text{Linear}_{v'}(\mathbf{F}_{\mathbf{T}})$. We use the [CLS] token in the output feature of $L$-th transformer layer for computing losses and retrieval, \emph{i.e.,} $\mathbf{F} = \mathbf{f}_L^{\text{CLS}}$, where $\mathbf{F}_L=(\mathbf{f}_L^{\text{CLS}}, \mathbf{f}_L^{1}, \mathbf{f}_L^{2}, ..., \mathbf{f}_L^{N})$ and $N$ is patch number of query images.

\begin{figure}
    \centering
    \includegraphics[width=\linewidth]{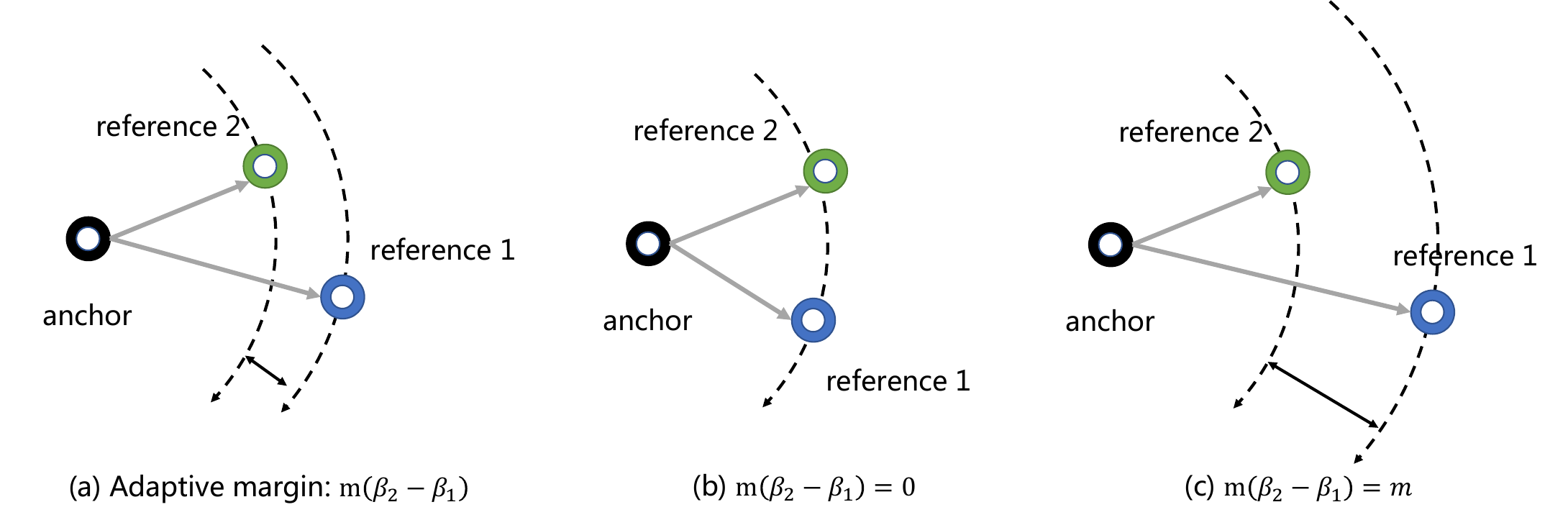}
    \caption{Illustration of adaptive triplet loss. Unlike the traditional triplet loss where the margin is fixed, the margin in our adaptive triplet loss is defined by the instruction similarity for the two query-instruction pairs that describe the same person. The features associated with similar instructions are pulled to be closer.}
    \label{fig:my_label}
\end{figure}

\subsubsection{Adaptive Triplet Loss} \label{sec:adaptive_triplet_loss}

Unlike typical triplet loss that defines positive and negative samples solely based on identities, instruct-ReID requires distinguishing images with different instructions for the same identity. Intuitively, an adaptive margin should be set to push or pull samples based on the instruction difference. Let ($\mathbf{F}^{a}_{i}$, $\mathbf{F}^{r_{1}}_{i}$, $\mathbf{F}^{r_{2}}_{i}$) be the $i$-th triplet in the current mini-batch, where $\mathbf{F}^{a}_{i}$ is an anchor sample, $\mathbf{F}^{r_{1}}_{i}$ and $\mathbf{F}^{r_{2}}_{i}$ are reference samples. We propose an adaptive triplet loss as  
% \begin{equation} \small
%     \begin{split}
%     \mathbf{\mathcal{L}}_{atri}=\frac{1}{N_{tri\dagger} } \sum_{i=1}^{N_{tri\dagger}}  \left \{ \mathbf{Sign}(\beta_1 - \beta_2) \left [  d(\mathbf{F}^{a}_{i},\mathbf{F}^{r_{1}}_{i}) + \\
%     \left ( \beta_1 - \beta_2  \right )m - d(\mathbf{F}^{a}_{i},\mathbf{F}^{r_{2}}_{i}) \right ] \right \}_{+} 
%     \end{split}
% \end{equation}
% \begin{equation} \small
%     \begin{split}
%     \mathbf{\mathcal{L}}_{atri}=\frac{1}{N_{tri\dagger} } \sum_{i=1}^{N_{tri\dagger}}  \left \{ \mathbf{Sign}(\beta_1 - \beta_2) \left [ \right \right d(\mathbf{F}^{a}_{i},\mathbf{F}^{r_{1}}_{i}) + \\ \left ( \beta_1 - \beta_2  \right \left \left)m - d(\mathbf{F}^{a}_{i},\mathbf{F}^{r_{2}}_{i}) \right ] \right \}_{+} 
%     \end{split}
% \end{equation}
\begin{equation} \small
    \begin{split}
    \mathbf{\mathcal{L}}_{atri}=\frac{1}{N_{tri\dagger} } \sum_{i=1}^{N_{tri\dagger}}  \{ \mathbf{Sign}(\beta_1 - \beta_2)  [  d(\mathbf{F}^{a}_{i},\mathbf{F}^{r_{1}}_{i}) + \\ \left ( \beta_1 - \beta_2  \right )m - d(\mathbf{F}^{a}_{i},\mathbf{F}^{r_{2}}_{i})  ]  \}_{+} 
    \end{split}
\end{equation}
where $N_{tri\dagger}$ and $m$ denote the number of triplets and a hyper-parameter for the maximal margin, respectively. $d$ is a Euclidean distance function, \emph{i.e.}, $d(\mathbf{F}^{a}_{i},\mathbf{F}^{r}_{i})= \left |\left | \mathbf{F}^{a}_{i} - \mathbf{F}^{r}_{i} \right | \right |_{2} ^{2}$. $\beta_1$ and $\beta_2$ are relatednesses between the anchor image and the corresponding reference image that consider the identity consistency and the instruction similarity for the adaptive margin. Mathematically, they can be defined as
\begin{equation} \small
    \beta_j = \mathbb{I}\left(y_a=y_{r_j}\right) \mathbf{Cos}\left \langle \mathbf{F}_{\mathbf{T}}^{a}, \mathbf{F}_{\mathbf{T}}^{r_{j}} \right \rangle,
\end{equation}
where $y_a$ and $y_{r_j}$ are the identity labels of the anchor image and the reference image, $\mathbb{I}(\cdot)$ is the indicator function, and $j=\{1, 2\}$ denotes the index of reference samples.

The concept of adaptive triplet loss is described by \textcolor{black}{Fig.~\ref{fig:my_label} (a)}. We discuss the adaptive loss in two cases. First, as shown in \textcolor{black}{Fig.~\ref{fig:my_label} (b)}, the margin is set to zero if the triplet has the same identity and the instructions of the two reference samples are equally similar to the instruction of the anchor sample. This makes the distances between the anchor and two reference points the same. Second, as shown in \textcolor{black}{Fig.~\ref{fig:my_label} (c)}, when there is a significant difference in instruction similarities, the margin distance between the anchor and two references becomes closer to the maximum value $m$, forcing the model to learn discriminative features. Adaptive triplet loss makes features from the same person become distinctive based on the similarity of instructions, which helps to retrieve images that align the requirements of given instructions in the CTCC-ReID and LI-ReID tasks.

\begin{figure*}[htbp]
    \centering
    \includegraphics[width=0.95\linewidth]{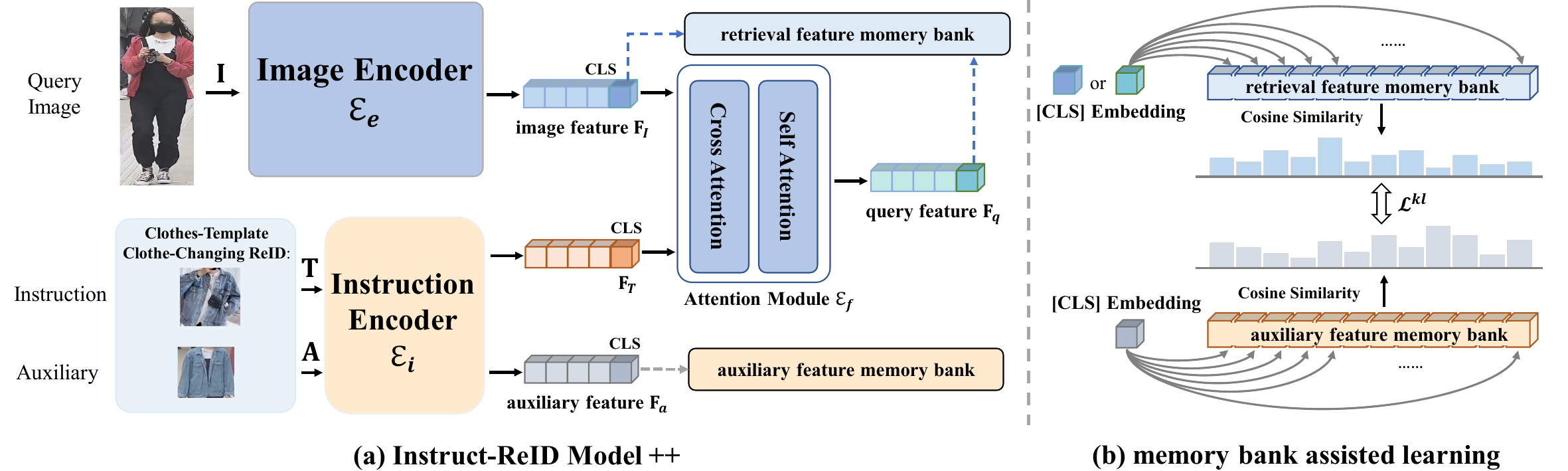}
    \caption{(a) The structure of the instruct-ReID++ Model. Auxiliary is employed only in training step, to endow model with powerful supervision (b) We propose a novel learning scheme, using two memory banks to guiding feature alignment.}
    \label{fig:IRMv2_method}
\end{figure*}

\subsubsection{Overall Loss Function} \label{sec:overall_loss} 

We impose an identification loss $\mathcal{L}_{id}$, which is the classification loss on identities, and an adaptive triplet loss $\mathcal{L}_{atri}$ on $\mathbf{F}$ and the final fusion features $\mathbf{F}_\mathbf{out}$ to supervise the model for training Trad-ReID, CC-ReID, VI-ReID, CTCC-ReID, and LI-ReID tasks. The overall loss is combined as
% \vspace{-0.5em}
\begin{equation} \small
    \begin{split}
    \mathbf{\mathcal{L}}= \alpha \mathcal{L}_{atri}(F) + \mathcal{L}_{id}(F) + \alpha \mathcal{L}_{atri}(F_{out}) + \mathcal{L}_{id}(F_{out})
    \end{split}
\end{equation}
% \vspace{-0.5em}
where $\mathbf{F}$ and $\mathbf{F}_{out}$ indicate the source of features used in calculating the loss and $\alpha$ is a hyperparameter to adjust the weight of different losses.

For the T2I-ReID task, we adopt a contrastive loss $\mathcal{L}_{cl}$ to align the image features $\mathbf{F}$ and text features $\mathbf{F}_\mathbf{T}$ to enable text-based person retrieval. We also employ a binary classification loss $\mathcal{L}_{match}$ to learn whether an inputted image-text pair is positive or negative, defined as:
\begin{equation} \small
    \begin{split}
    \mathbf{\mathcal{L}}=\mathcal{L}_{cl}({F}) + \mathcal{L}_{match}({F_{out}})
    \end{split}
\end{equation}
where, $\mathcal{L}_{match}({F_{out}}) = \mathcal{C}_{e} (\hat{y},F_{out}) = \mathcal{C}_{e} [\hat{y},\mathcal{E}_f(\mathbf{F}, \mathbf{F_{T}})] $, respectively. $\mathcal{C}_{e}$ represents a binary cross-entropy function, $\hat{y}$ is a 2-dimension one-hot vector representing the ground-truth label \emph{i.e.,} $[0, 1]^\intercal$ for the positive pair and $[1, 0]^\intercal$ for the negative pair, which is formed by matching text features with corresponding image features before inputting into the attention module $\mathcal{E}_f$.

\subsection{Model Architecture of IRM++} \label{sec:IRM++}
The overview of the proposed IRM++ is shown in Fig.~\ref{fig:IRMv2_method} (a), which employs a pretrained ViT encoder $\mathcal{E}_{e}$ for image encoding and utilizes a vision-language model, \emph{e.g.,} ALBEF~\cite{li2021align}, as the instruction encoder $\mathcal{E}_{i}$. Our IRM++ processes the outputs of the query and gallery features in two basic formats. For the \textit{query feature}, the query image $\mathbf{I}$ and the paired instruction $\mathbf{T}$ are separately input to the image encoder and instruction encoder to extract image features $\mathbf{F}_{I}=\mathcal{E}_{e}(\mathbf{I})$ and instruction features $\mathbf{F}_{T}=\mathcal{E}_{i}(\mathbf{T})$. These features are then combined by an attention module $\mathcal{E}_{f}$ to integrate the information from the query image and retrieval intent instruction, ultimately outputting the query feature $\mathbf{F}_{q}=\mathcal{E}_{f}(\mathbf{F}_{I},\mathbf{F}_{T})$. For the \textit{gallery feature}, gallery images from various ReID tasks are input to the image encoder of IRM++, and the extracted image features are used as the gallery features for retrieval, \emph{e.g.,} $\mathbf{F}_{g}=\mathbf{F}_{I}=\mathcal{E}_{e}(\mathbf{I})$. IRM++ removes the task-specific instruction inputs and provides a more general way of extracting gallery features, which is suitable for the task-free evaluation setting. For clarity, in the subsequent discussion, we assign both the query feature $\mathbf{F}_{q}$ and the gallery feature $\mathbf{F}_{g}$ to the retrieval feature $\mathbf{F}_{r}$.

% Task-free evaluation setting requires the non-differentiation of gallery image data across various ReID tasks. In the inference phase, the gallery image features of different ReID tasks are extracted from the same positions and lack specific instruction commands.
% \subsubsection{Memory Bank Assisted Model Learning} \label{sec:memory bank}
To facilitate IRM++ in the task-free setting, we utilize two memory banks, shown in Fig.~\ref{fig:IRMv2_method} (b), and implement a contrastive learning method. In the training phase, we utilize the instruction annotation as the auxiliary data $\mathbf{A}$ and the auxiliary feature $\mathbf{F}_{a}=\mathcal{E}_{i}(\mathbf{A})$ is derived from the instruction encoder. We designate images in the training set as query type or gallery type. For instance, in CTCC-ReID, images paired with template clothes image instructions are denoted as query type, while images wearing target clothes are classified as gallery type. Similarly, in LI-ReID, images paired with language description instructions are designated as query type, and those that match the retrieval target are classified as gallery type. We extract the query feature $\mathbf{F}_{q}$ or the gallery feature $\mathbf{F}_{g}$ based on the corresponding query or gallery type of the training set images, and assign them as the retrieval feature $\mathbf{F}_{r}$.
% As the ideal $\mathbf{F}_{r}$ have the ability to sense the modification of instructions, the similarity of different $\mathbf{F}_{r}$ has positive correlation to the similarity of different instructions. However, gallery input in instruct-ReID is single image with no instruction, so we replace the similarity of instructions with the similarity of auxiliaries. $\mathbf{F}_{r}$ and $\mathbf{F}_{a}$ are stored in different branch memory banks. 
Based on the retrieval feature $\mathbf{F}_{r}$ and auxiliary feature $\mathbf{F}_{a}$, we maintain the retrieval feature memory bank $\mathbf{M}_{r}=\{m^{r}_{1}, m^{r}_{2},...,m^{r}_{N}\}$ and auxiliary feature memory bank $\mathbf{M}_{a}=\{m^{a}_{1}, m^{a}_{2},...,m^{a}_{N}\}$, where N represents the number of identities in the training set. Both ranks are initialized randomly, during the training process, the retrieval feature $\mathbf{F}_{r}$ associated with identity $i$ will be stored in $m^{r}_{i}$, replacing the original feature inside it. Similarly, the feature $m^{a}_{i}$ will be replaced by auxiliary feature $\mathbf{F}_{a}$ that associates with identity $i$. For each iteration, we compute the cosine similarity of $\mathbf{F}_{r}$ with each element in $\mathbf{M}_{r}$ and obtain the similarity score $\mathbf{S}_{r}=\text{Similarity}(\mathbf{F}_{r}, \mathbf{M}_{r})=\{s^{r}_{1}, s^{r}_{2},...,s^{r}_{N}\}$. The calculation process for each specific $s^{r}_{i}$ can be defined as
\begin{equation} \small
    s^{r}_{i} = \mathbf{Cos}\left \langle \mathbf{F_{r}}, \mathbf{m^{r}_{i}} \right \rangle.
\end{equation}
We do the same way to the similarity score of auxiliary, getting $\mathbf{S}_{a}=\text{Similarity}(\mathbf{F}_{a}, \mathbf{M}_{a})=\{s^{a}_{1}, s^{a}_{2},...,s^{a}_{N}\}$. 
In this way, the similarity score of the auxiliary can be seen as a soft label, while the ground truth identity can be seen as a hard label. We utilize a contrastive learning loss to guide the learning with soft label, which is defined as
\begin{equation} \small
    \begin{split}
    \mathbf{\mathcal{L}}_{soft}=\mathbf{\mathcal{L}}_{kl}(\mathbf{S}_{a}, \mathbf{S}_{r}) =\frac{1}{N_{kl} } \sum_{i=1}^{N_{kl}} \mathcal{P}(s^{a}_{i})  \mathbf{log} \frac{\mathcal{P}(s^{r}_{i})}{\mathcal{P}(s^{a}_{i}) + \xi },
    % \mathbf{Prob}(s^{a}_{i}) /  d(\mathbf{F}^{a}_{i},\mathbf{F}^{r_{1}}_{i}) + \\ \left ( \beta_1 - \beta_2  \right )m - d(\mathbf{F}^{a}_{i},\mathbf{F}^{r_{2}}_{i})  ]  \}_{+} 
    \end{split}
\end{equation}
% \begin{equation} \small
%     \begin{split}
%     \mathbf{\mathcal{L}_{soft}}=\mathbf{KL}(\mathbf{S}_{a}||\mathbf{S}_{r})
%     \end{split}
% \end{equation}
where $N_{kl}$ and $\xi$ denote the number of contrastive learning samples and a smoothing factor to prevent division by zero, respectively. $\mathcal{P}(\cdot)$ is a softmax function. We turned ground truth into a $\mathbf{N}$-length one-hot label $\mathbf{S}_{gt}=\{s^{gt}_{1},s^{gt}_{2},...,s^{gt}_{N}\}$, with the same shape to the similarity score. $s^{gt}_{i}=1$ if $\mathbf{F}_{r}$ with identity $i$, instead $s^{gt}_{i}=0$. We also use contrastive loss to guide the supervision of this hard label and similarly by
\begin{equation} \small
    \begin{split}
    \mathbf{\mathcal{L}}_{hard}=\mathbf{\mathcal{L}}_{kl}(\mathbf{S}_{gt}, \mathbf{S}_{r}) =\frac{1}{N_{kl} } \sum_{i=1}^{N_{kl}} \mathcal{P}(s^{{gt}}_{i})  \mathbf{log} \frac{\mathcal{P}(s^{{r}}_{i})}{\mathcal{P}(s^{{gt}}_{i}) + \xi }.
    % \mathbf{Prob}(s^{a}_{i}) /  d(\mathbf{F}^{a}_{i},\mathbf{F}^{r_{1}}_{i}) + \\ \left ( \beta_1 - \beta_2  \right )m - d(\mathbf{F}^{a}_{i},\mathbf{F}^{r_{2}}_{i})  ]  \}_{+} 
    \end{split}
\end{equation}
% \begin{equation} \small
%     \begin{split}
%     \mathbf{\mathcal{L}_{hard}}=\mathbf{KL}(\mathbf{S}_{gt}||\mathbf{S}_{r})
%     \end{split}
% \end{equation}
% description, guiding the model to distinguish various query instructions. It helps the model sense the diversity of input. On the other hand, 
The soft label provides detailed and fine-grained instruction supervision and the hard label ensures the model learns reliable identity information. By combining the contrastive loss and the identification loss $\mathcal{L}_{id}$, the overall loss is defined as
\begin{equation} \small
    \begin{split}
    \mathbf{\mathcal{L}}=\mathbf{\mathcal{L}}_{id}(F) + \mathbf{\mathcal{L}}_{soft}(F)+\beta\times\mathbf{\mathcal{L}}_{hard}(F),
    \end{split}
\end{equation}
where $\beta$ is a hyperparameter to adjust the loss weights.

\section{Experiments}
\subsection{Datasets and Evaluation Metric}
\subsubsection{Training datasets} \label{sec:traning_datasets}
% Describe the categories of datasets and why we choose these datasets for training, briefly introduce typeciall datasets and rest in sup.
Using the proposed \textbf{OmniReID++} benchmark, we train IRM and IRM++ under two training settings:
(1) Single-task Learning (STL): The model is trained separately on each dataset corresponding to the individual task.
(2) Multi-task Learning (MTL): The model is trained jointly using all datasets included in OmniReID++.
\subsubsection{Evaluation Datasets}
% \label{sec:test_datasets}
To assess the efficacy of our proposed method across diverse ReID scenarios, we conduct the following tests according to the OmniReID++ benchmark settings: 
(1) task-specific evaluation setting: we independently evaluate the performance of IRM on each test dataset included in the OmniReID++ benchmark. We then compare our results with existing state-of-the-art methods.
(2) task-free evaluation setting: we consolidate all test datasets from the OmniReID++ benchmark into a unified test set and present the evaluation results of IRM++ under this configuration. Additionally, we assess the performance of IRM++ on each sub-test dataset to compare with state-of-the-art methods. We refer to sub-test set and joint test set evaluations as \underline{\emph{Single Test Set Evaluation}} and \underline{\emph{Joint Test Set Evaluation}}, respectively.

\subsubsection{Evaluation Metric}

In the single test set evaluation setting, we use metrics such as the Cumulative Matching Characteristic (CMC) curve at top-1 and mAP (mean Average Precision) to assess ReID performances quantitatively. For the joint test set evaluation setting, in addition to the mAP metric, we also utilize our proposed mAP$\tau$ evaluation metric to validate the performance of the metrics at different thresholds.

\subsection{Implementation Details}
For the editing transformer, we use the plain ViT-Base with the ReID-specific pretraining~\cite{zhu2022pass}. We adopt ALBEF~\cite{li2021align} as our instruction encoder. All images are resized into 256$\times$128 for training and testing. We use the AdamW optimizer with a base learning rate of 1e-5 and a weight decay of 5e-4. We linearly warmup the learning rate from 1e-7 to 1e-5 for the first 1000 iterations. Random cropping, flipping, and erasing are used for data augmentation during training. For each training batch, we randomly select 32 identities with 4 image samples for each identity. The Single-task Learning takes one NVIDIA V100 GPU and the Multi-task Learning takes 32 NVIDIA V100 GPUS for training, respectively.

% \noindent \textbf{IRM++.} We follow the implementation details of IRM and also use the plain ViT-Base pretrained on ReID as our image encoder, and take ALBEF as our instruction encoder. 

\begin{table}[t]
  \footnotesize
  \centering
  \renewcommand\arraystretch{1.2}
  \caption{The performance of Clothse-Changing ReID of our method and the state-of-the-art methods. Mean average precision (mAP) and Top1 are used to quantify the accuracy. \dag~denotes that the model is trained with multiple datasets. *~denotes that the model is pre-trained on 4 million images.\textcolor{black}{All methods are evaluated using the rank-1 and mAP metrics.}}
    \begin{tabular}{l|cc|cc|cc}
    \hline
    \multicolumn{1}{l|}{\multirow{2}{*}{\textbf{Methods}}}&\multicolumn{2}{c|}{\textbf{LTCC}} &\multicolumn{2}{c|}{\textbf{PRCC}} &\multicolumn{2}{c}{\textbf{VC-Clothes}} \\
    \cline{2-7} & mAP & Top1 & mAP & Top1 & mAP & Top1\\
    \hline
    HACNN~\citep{li2018harmonious} & 26.7 & 60.2 & - & 21.8 & - & -\\
    RGA-SC~\citep{zhang2020relation} & 27.5 & 65.0 & - & 42.3 & 67.4 & 71.1\\
    PCB~\citep{sun2018beyond} & 30.6 & 65.1 & 38.7 & 41.8 & 62.2 & 62.0\\
    IANet~\citep{hou2019interaction} & 31.0 & 63.7 & 45.9 & 46.3 & - & -\\
    CAL~\citep{gu2022clothes} & 40.8 & 74.2 & - & - & - & -\\
    TransReID~\citep{he2021transreid} & - & - & - & 44.2 & 71.8 & 72.0\\
    \hline
    IRM (STL) & 46.7 & 66.7 & 46.0 & 48.1 & \textbf{80.1} & \textbf{90.1} \\
    IRM (MTL)\dag & \textbf{52.0} & \textbf{75.8} & \textbf{52.3} & \textbf{54.2} & 78.9 & 89.7 \\
    \hline
    \end{tabular}%
  \label{tab:CC-ReID}%
\end{table}%

\begin{table}[t]\footnotesize
  \centering
  \renewcommand\arraystretch{1.2}
  \caption{Comparison with the state-of-the-art methods on visible-infrared ReID and text-to-image ReID. The VI-ReID setting is \textit{VIS-to-IR} and \textit{IR-to-VIS} in LLCM. \dag~denotes the model is trained with multiple datasets.*~denotes the model is trained under the same image shape as IRM, \emph{i.e.,} 256$\times$128. \textcolor{black}{All methods are evaluated using the rank-1 and mAP metrics.}}
    \begin{tabular}{l|cc|cccc}
    \hline
    \multicolumn{1}{l|}{\multirow{3}{*}{\textbf{Methods}}} &\multicolumn{2}{c|}{\textbf{T2I-ReID}}&\multicolumn{4}{c}{\textbf{VI-ReID: LLCM}} 
    \\
    \cline{2-7} 
    &\multicolumn{2}{c|}{\textbf{CUHK-PEDES}} & \multicolumn{2}{c}{\textbf{VIS-to-IR}} &\multicolumn{2}{c}{\textbf{IR-to-VIS}}  \\
    & mAP & Top1 & mAP & Top1 & mAP & Top1 \\
    \hline
    ALBEF~\cite{li2021align} & 56.7 & 60.3&- & -&-&- \\
    CAJ~\cite{ye2021channel} & - & - & 59.8 & 56.5 & 56.6 & 48.8\\
    MMN~\cite{zhang2021towards} & - & - & 62.7 & 59.9 & 58.9 & 52.5\\
    DEEN~\cite{zhang2023diverse} & - & - & 65.8 & 62.5 & 62.9 & 54.9\\
    SAF~\cite{li2022learning} & 58.6& 64.1& - & -&-&-\\
    PSLD~\cite{han2021text}& 60.1 & 64.1&-&-&-&- \\
    RaSa*~\cite{bai2023rasa}& 63.9 & 69.6 &-&-&-&- \\
    \hline
    IRM (STL) & 65.3 & 72.8 & 66.6 & 66.2 & 64.5 & 64.9 \\
    IRM (MTL)\dag & \textbf{66.5} & \textbf{74.2} & \textbf{67.5} & \textbf{66.7} & \textbf{67.2} & \textbf{65.7} \\
    \hline
    \end{tabular}%
  \label{tab:t2i_cross}%
\end{table}%

\subsection{Task-specific Evaluation Setting Results}

In this subsection, we validate the effectiveness of the proposed IRM under the task-specific setting. All results are obtained based on the OmniReID++ benchmark, and we compare with current state-of-the-art methods on the six tasks, using commonly used metrics such as mAP and CMC (rank1)

\noindent \textbf{Clothes-Changing ReID (CC-ReID).}
As shown in Tab.~\ref{tab:CC-ReID}, IRM outperforms all state-of-the-art methods on LTCC, PRCC and VC-Clothes, showing that the model can effectively extract clothes-invariant features following the instructions, \emph{e.g.,} ``Ignore clothes''. Specifically, on STL, IRM improves CAL~\cite{gu2022clothes}, TransReID~\cite{he2021transreid} by \textbf{+5.9\%} mAP and \textbf{+8.3\%} mAP on LTCC and VC-Clothes, respectively. Using multi-task training further improves the performance of IRM to \textbf{52.0\%} mAP on LTCC, and reaches a new state-of-the-art on PRCC with a \textbf{52.3\%} mAP. While multi-task learning leads to slightly lower performance than single-task learning on VC-Clothes, IRM still achieves a higher Top-1 than TransReID. We analyze that this drop is due to the domain gap between VC-Clothes (Synthetic) and datasets (Real) and leave it for future work.

\noindent \textbf{Clothes Template Based Clothes-Changing ReID (CTCC-ReID).}
Our method achieves desirable performance on the CTCC-ReID task in Tab.~\ref{tab:CT-ReID}, which shows that a fixed instruction encoder is enough for this tough task. Concretely, when only trained on COCAS+ Real1, IRM outperforms BC-Net~\cite{yu2020cocas} and DualBCT-Net~\cite{li2022cocas+}, both of which learn an independent clothes branch, by \textbf{+9.6\%} mAP and \textbf{+2.2\%} mAP, respectively. By integrating the knowledge on other instruct ReID tasks during multi-task learning, we are able to further improve the performance of IRM, achieving an mAP of \textbf{41.7\%} and pushing the performance limits on CTCC-ReID.

\begin{table*}[t]\small
  \centering
  \renewcommand\arraystretch{1.1}
  \caption{Performance comparison with the state-of-the-art methods on Clothes
  -Template Clothes-Changing ReID, Language-Instructed ReID, and Traditional ReID. \dag~denotes that the model is trained with multiple datasets. *~denotes that the model is pretrained on 4 million pedestrian images. \underline{The size of the input images used in the table is 256x128.} \textcolor{black}{All methods use rank-1 and mAP metrics.}}
    \begin{tabular}{l|cc|cc|cccccc}
    \hline
    \multicolumn{1}{l|}{\multirow{3}{*}{\textbf{Methods}}} &\multicolumn{2}{c|}{\textbf{CTCC-ReID}}&\multicolumn{2}{c|}{\textbf{LI-ReID}} &\multicolumn{6}{c}{\textbf{Trad-ReID}} 
    \\
    \cline{2-11} 
    &\multicolumn{2}{c|}{\textbf{COCAS+ Real2}}&\multicolumn{2}{c|}{\textbf{COCAS+ Real2}} &\multicolumn{2}{c}{\textbf{Market1501}} &\multicolumn{2}{c}{\textbf{MSMT17}} &\multicolumn{2}{c}{\textbf{CUHK03}}  \\
    & mAP & Top1& mAP & Top1& mAP & Top1& mAP & Top1& mAP & Top1 \\
    \hline
    Baseline & - & -&14.9&31.6& -& -& -& -& -& -\\
    TransReID~\cite{he2021transreid} & 5.5 & 17.5&-&-&86.8 & 94.4 & 61.0 & 81.8 & - & -\\
    % Pixel Sampling~\cite{shu2021semantic} & 2.1 & 11.6&-&-& -& -& -& -& -& -\\
    BC-Net~\cite{yu2020cocas} & 22.6 & 36.9&-&-& -& -& -& -& -& -\\
    DualBCT-Net~\cite{li2022cocas+} & 30.0 & 48.9&-&-& -& -& -& -& -& -\\
    SAN~\cite{jin2020semantics}&-&-&-&-&
    88.0 & 96.1 & - & - & 76.4 & 80.1
    \\
    HumanBench\dag~\cite{tang2023humanbench}&-&-&-&-&89.5 & - & 69.1 & - & 77.7 & -
    \\
    PASS*~\cite{zhu2022pass}&-&-&-&-&93.0 & 96.8 & 71.8 & 88.2 & - & - 
    \\
    \hline
    IRM (STL) & 32.2 & 54.8 & 30.7 & 60.8 & 92.3 & 96.2 & 71.9 & 86.2 & 83.3 & 86.5 \\
    IRM (MTL)\dag & \textbf{41.7} & \textbf{64.9} & \textbf{39.8} & \textbf{71.6} & \textbf{93.5} & \textbf{96.5} & \textbf{72.4} & \textbf{86.9} & \textbf{85.4} & \textbf{86.5} \\
    \hline
    \end{tabular}%
  \label{tab:CT-ReID}%
\end{table*}%

\noindent \textbf{Visible-Infrared ReID (VI-ReID).} We validate the performance of IRM on Visible-Infrared ReID datasets LLCM, which is a new and challenging low-light
cross-modality dataset and has a more significant number
of identities and images. From Tab.~\ref{tab:t2i_cross}, we can see that the results on the two test modes show that the proposed IRM achieves competitive performance against all other state-of-the-art methods. Specifically, for the VIS-to-IR mode on LLCM, IRM achieves \textbf{67.5\%} mAP and exceeds previous state-of-the-art methods like DEEN~\cite{zhang2023diverse} by \textbf{+1.7\%}. For the IR-to-VIS mode on LLCM, IRM achieves \textbf{65.7\%} Rank-1 accuracy
and \textbf{67.2\%} mAP, which is a new state-of-the-art result. The results validate the effectiveness of our method.

\noindent \textbf{Text-to-Image ReID (T2I-ReID).} As shown in Tab.~\ref{tab:t2i_cross}, IRM shows competitive performance with a mAP of \textbf{66.5\%} on the CUHK-PEDES~\cite{li2017person}, which is \textbf{+2.6\%}, \textbf{+6.4\%}, \textbf{+7.9\%} higher than previous methods  RaSa~\cite{bai2023rasa} (63.9\%), PSLD~\cite{han2021text} (60.1\%), SAF~\cite{li2022learning} (58.6\%). Because a few images in CUHK-PEDES training set are from the test sets of Market1501 and CUHK03, we filtered out duplicate images from the test sets during the multi-task learning (MTL) process. Following the common research works, the testing is conducted on uniformly resized images with a resolution of 256$\times$128.

\noindent \textbf{Language-Instructed ReID (LI-ReID).}
As a novel setting, no previous works have been done to retrieve a person using several sentences as the instruction, therefore, we compare IRM with a straightforward baseline. In the baseline method, only a ViT-Base is trained on COCAS+ Real1 images without utilizing the information from language instruction, leading to poor person re-identification ability. As shown in Tab.~\ref{tab:CT-ReID}, IRM improves the baseline by \textbf{+15.8\%} mAP, because IRM can integrate instruction information into identity features. With MTL, IRM achieves extra \textbf{+9.1\%} performance gain by using more images and general information in diverse ReID tasks.

\noindent \textbf{Traditional ReID (Trad-ReID).} 
IRM also shows its effectiveness on Trad-ReID tasks in Tab.~\ref{tab:CT-ReID}. Specifically, when trained on a single dataset, compared with PASS~\cite{zhu2022pass}, IRM achieves comparable performance on Market1501, MSMT17 and \textbf{+6.9\%} performance gain on CUHK03 compared with SAN~\cite{jin2020semantics}. With multi-task training, IRM can outperform the recent multi-task pretraining HumanBench~\cite{tang2023humanbench} and self-supervised pretraining~\cite{zhu2022pass}. We do not compare with SOLDIER~\cite{Chen_2023_CVPR} because it only reports ReID results with the image size of 384$\times$192 instead of 256$\times$128 in our method. More importantly, SOLDIER primarily focuses on pretraining and is evaluated only on Trad-ReID, while the claimed contribution of IRM is to tackle multiple ReID tasks with one model.

\begin{table*}[t]
%\resizebox{\textwidth}{!}{
  \centering
  \scriptsize
  \renewcommand\arraystretch{1.3}
  \caption{\textcolor{black}{The performance comparison of IRM++ with the state-of-the-art methods on 6 instruct-ReID tasks and all results are presented based on task-free evaluation setting. \dag~denotes that the test mode is VIS-to-IR and \ddag~denotes IR-to-VIS mode on LLCM. *~denotes that the model utilizes the same network architecture and loss function as IRM but outputs the gallery features of different tasks from a unified location as a baseline method. The size of the input images used in the table is 256x128. All methods are evaluated using the mAP metric.}}
    \begin{tabular}{l|c|c|c|cc|ccc|ccc}
    \hline
    \multicolumn{1}{l|}{\multirow{2}{*}{\textbf{Methods}}}&\textbf{CTCC-ReID}&\textbf{LI-ReID}&\textbf{T2I-ReID}&\multicolumn{2}{c|}{\textbf{VI-ReID}}&\multicolumn{3}{c|}{\textbf{CC-ReID}} &\multicolumn{3}{c}{\textbf{Trad-ReID}} \\
    \cline{2-12} & Real2 & Real2 & CUHK. & LLCM\dag & LLCM\ddag & LTCC & PRCC & VC-Clo. & Market1501 & MSMT17 & CUHK03 \\
    \hline
    SOTA & 30.0~\cite{li2022cocas+} & 14.9 & 63.9~\cite{bai2023rasa} & 62.9~\cite{zhang2023diverse} & 65.8~\cite{zhang2023diverse}  & 40.8~\citep{gu2022clothes} & 45.9~\citep{hou2019interaction} & 71.8~\citep{he2021transreid} & 93.0~\cite{zhu2022pass} & 71.8~\cite{zhu2022pass} & 77.7~\cite{tang2023humanbench} \\
    \hline
    \textcolor{black}{IRM* (STL)} & 22.2 & 28.7 & 64.2 & 64.6 & 66.5 & 46.1 & 46.3 & \textbf{80.4} & 92.4 & 72.0 & 83.5 \\
    \textcolor{black}{IRM* (MTL)} & 26.3 & 31.5 & 66.1 & 66.0 & 66.8 & 46.6 & 47.3 & 78.2 & 93.2 & 72.4 & 84.6 \\
    \hline
    IRM++ (STL) & 32.2 & 35.8 & 64.3 & 64.5 & 66.3 & 46.2 & 46.5 & 80.3 & 92.1 & 72.2 & 83.3 \\
   IRM++ (MTL) & \textbf{35.3} & \textbf{37.3} & \textbf{66.4} & \textbf{66.1} & \textbf{67.1} & \textbf{46.7} & \textbf{47.7} & 78.4 & \textbf{93.3} & \textbf{72.5} & \textbf{84.7} \\
    \hline
    \end{tabular}%
  \label{tab:task-free single}%
\end{table*}%

% \subsection{Results on Instruct-ReID-G WITH new metric}

% \vspace{-0.5cm}
\begin{table}[ht]
%\resizebox{\textwidth}{!}{
  \centering
  \scriptsize
  \renewcommand\arraystretch{1.2}
  \caption{The performance of Joint Test Set Evaluation Setting results includes the $mAP\tau$ results at three thresholds. *~denotes that the model is trained by triplet loss as a baseline method.}
    \resizebox{0.45\textwidth}{!}{\begin{tabular}{cc|c|c|c}
    % \toprule
    \hline
    % \multicolumn{2}{c|}{\multirow{2}{*}{\textbf{IRM}}}&\textbf{CTCC-ReID}&\textbf{LI-ReID}&\textbf{T2I-ReID} \\
    % \cline{3-5} & & Real2 & Real2 & CUHK. \\
    Method&mAP&mAP$_{25}$&mAP$_{50}$&mAP$_{75}$ \\
    % \cline{2-9} & mAP & mAP & mAP & mAP & mAP & mAP & mAP & mAP\\
    % \midrule
    \hline
    DualBCT-Net~\cite{li2022cocas+} & 37.1 & - & - & - \\
    RaSa~\cite{bai2023rasa} & 43.2 & - & - & - \\
    PASS~\cite{zhu2022pass} & 42.2 & - & - & - \\
    CAL~\citep{gu2022clothes} & 37.3 & - & - & - \\
    DEEN~\cite{zhang2023diverse} & 29.4 & - & - & - \\
    \hline
    % \midrule
    \textcolor{black}{IRM*} & 56.2 & 55.8 & 55.4 & 54.2 \\
    IRM++ & \textbf{65.3} & \textbf{64.7} & \textbf{62.1} & \textbf{60.5} \\
    % \bottomrule
    \hline
    \end{tabular}}
  \label{tab:task-free joint}%
\end{table}%

\subsection{Task-free Evaluation Setting Results}
In this subsection, we validate the effectiveness of the proposed IRM++ under the task-free setting. All results are obtained based on the OmniReID++ benchmark. Specifically, we first evaluate the performance of IRM++ on the six tasks in the Single Test Set Evaluation Setting, using metrics such as mAP and CMC (rank1). Then, we provide the mAP$\tau$ results of IRM++ and current state-of-the-art methods under the Joint Test Set Evaluation Setting for reference, aiming to facilitate research in this field.

\noindent \textbf{Single Test Set Evaluation.} 
We evaluate the performance of IRM++ on each single test set, which is a subset of the task-free evaluation setting. As shown in Tab.~\ref{tab:task-free single}, for clothes template based clothes-changing ReID (CTCC-ReID), language-instructed ReID (LI-ReID), clothes-changing ReID (CC-ReID), visible-infrared ReID (VI-ReID), text-to-image ReID (T2I-ReID), and traditional ReID (Trad-ReID), under Multi-task learning (MTL) training, our IRM++ achieves state-of-the-art (SOTA) results, surpassing the best-performing methods by \textbf{+0.3\%} to \textbf{+22.4\%} mAP. Experimental results demonstrate that IRM++ effectively extracts generic gallery features and compares them with features extracted from query images and instructions for accurate retrieval. 
Besides, we conduct a baseline method \textcolor{black}{IRM*} by employing a triplet loss as the supervision function. For CTCC-ReID and LI-ReID tasks, IRM++ surpasses the baseline method by \textbf{+10.0\%} and \textbf{+9.0\%} mAP (STL), \textbf{+7.1\%} and \textbf{+5.8\%} mAP (MTL), which indicates that the memory bank assisted contrastive learning utilized in IRM++ further enhances the model performance compared to optimization with triplet loss on these two ReID tasks. We analyze that the memory bank-assisted contrastive loss employs multiple negative samples for each iteration, providing stronger supervision during training compared to triplet-based optimization methods, which only utilize one negative sample per iteration. For CC-ReID, VI-ReID, and Trad-ReID, our IRM++ and baseline method achieve similar results, with experimental variations within \textbf{0.3\%} mAP under Single-task learning (STL) training. 

\noindent \textbf{Joint Test Set Evaluation.} 
As shown in Tab.~\ref{tab:task-free joint}, we unify all testing set data into a joint set and provide benchmark testing results on the proposed dataset. We also perform inference with existing state-of-the-art methods on this testing set for comparison. Since DualBCT Net~\cite{li2022cocas+}, RaSa~\cite{bai2023rasa}, PASS~\cite{zhu2022pass}, CAL~\citep{gu2022clothes} and DEEN~\cite{zhang2023diverse} cannot receive multimodal instruction inputs, \emph{e.g.}, clothes template images and description texts, and the instruction similarity is unavailable, we only measure the traditional mAP metric on the Joint Test dataset. We also provide the inference performance of the baseline method \textcolor{black}{IRM*} on the newly proposed testing set. Under the traditional mAP evaluation metric, our IRM++ achieves the best result of \textbf{65.3\%} mAP, surpassing the baseline method and existing methods by \textbf{9.1\%} mAP and \textbf{22.1\%} mAP, respectively. The results demonstrate that IRM++ exhibits outstanding performance in the task-free evaluation setting. Based on instruction similarity and our proposed mAP$\tau$ evaluation metric, we provide evaluation results at three different thresholds of 0.25, 0.50, and 0.75. Compared to the baseline method, IRM++ achieves better results of \textbf{+8.9\%} (mAP$_{25}$), \textbf{+6.7\%} (mAP$_{50}$), and \textbf{+6.3\%} (mAP$_{75}$). We observe that under the mAP$\tau$ evaluation metric, the test results of our proposed IRM++ show a decreasing trend as the threshold increases. This indicates that with increasing thresholds, more gallery samples with the same identity as the query image but not matching the instruction attributes are corrected back to negative samples, aligning with our proposed mAP$\tau$ evaluation metric. The selection of thresholds for different models and scenarios is an area worthy of future research.

\subsection{Ablation Study}
% In this section, we explore the effectiveness of the editing transformer and  adaptive triplet loss for the ReID tasks by conducting ablation experiments.

\begin{table*}[t]
%\resizebox{\textwidth}{!}{
  \centering
  \scriptsize
  \renewcommand\arraystretch{1.3}
  \caption{Ablation study. The performance comparison of our editing transformer and using ViT base transformer (w/o editing), and the comparisons with triplet loss (w/o $\mathcal{L}_{atri}$) and the proposed adaptive triplet loss in terms of mAP. \dag~denotes that the test mode is VIS-to-IR and \ddag~denotes IR-to-VIS mode on LLCM.\textcolor{black}{All methods are evaluated using the mAP metric.}}
    \begin{tabular}{l|c|c|c|cc|ccc|ccc|c}
    \hline
    \multicolumn{1}{l|}{\multirow{2}{*}{\textbf{Methods}}}&\textbf{CTCC-ReID}&\textbf{LI-ReID}&\textbf{T2I-ReID}&\multicolumn{2}{c|}{\textbf{VI-ReID}}&\multicolumn{3}{c|}{\textbf{CC-ReID}} &\multicolumn{3}{c|}{\textbf{Trad-ReID}} & \multirow{2}{*}{\textbf{Avg.}} \\
    \cline{2-12} & Real2 & Real2 & CUHK. & LLCM\dag & LLCM\ddag & LTCC & PRCC & VC-Clo. & Market1501 & MSMT17 & CUHK03 & \\
    % \cline{2-9} & mAP & mAP & mAP & mAP & mAP & mAP & mAP & mAP\\
    \hline
    IRM (STL) & 32.2 & 30.7 & 65.3 & 66.6 & 64.5 & 46.7 & 46.0 & 80.1 & 92.3 & 71.9 & 83.3 & 61.8\\
    % avg 61.8 = (32.2+30.7+65.3+66.6+64.5+46.7+46.0+80.1+92.3+71.9+83.3)/11
    w/o editing & 32.6 & 30.5 & 65.7 & 66.2 & 65.1 & 46.2 & 45.7 & 77.8 & 92.5 & 71.4 & 83.2 & 61.5\\
    % avg 61.5 = (32.6+0.5+65.7+66.2+65.1+46.2+45.7+77.8+92.5+71.4+83.2)/11
    w/o $\mathcal{L}_{atri}$ & 31.5 & 30.2 & - & 66.6 & 64.5 & 46.7 & 46.0 & 80.1 & 92.3 & 71.9 & 83.3 & 61.3\\
    % avg 61.3 = (31.5+30.2+66.6+64.5+46.7+46.0+80.1+92.3+71.9+83.3)/10
     % fusion & 28.7 & 25.9 & 47.4 & 39.8 & 72.1 & 86.4 & 64.5 & 83.8\\
    \hline
    IRM (MTL) & 41.7 & 39.8 & 66.5 & 67.5 & 67.2 & 52.0 & 52.3 & 78.9 & 93.5 & 72.4 & 85.4 & 65.1\\
    % avg  65.1= (41.7+39.8+66.5+67.5+67.2+52.0+52.3+78.9+93.5+72.4+85.4)/11
   w/o editing & 41.0 & 38.8 & 66.1 & 67.3& 65.2 & 51.2 & 51.0 & 78.6 & 93.0 & 71.8 & 85.1 & 64.5\\
   % avg 64.5 = (41.0+38.8+66.1+67.3+65.2+51.2+51.0+78.6+93.0+71.8+85.1)/11
    w/o $\mathcal{L}_{atri}$ & 40.7 & 39.2 & 65.2 & 68.4 & 66.3 & 52.0 & 52.4 & 77.9 & 92.9 & 72.0 & 85.5 & 64.8\\
    % avg 64.8 = (40.7+39.2+65.2+68.4+66.3+52.0+52.4+77.9+92.9+72.0+85.5)/11
    % w/o adapter & 41.4 & - & - & - & - & - & - & -\\
    \hline
    \end{tabular}%
  \label{tab:ablation study}%
\end{table*}%

\noindent \textbf{Editing Transformer.}
To verify the effectiveness of the instruction integrating design in the editing transformer, we compare it with the traditional ViT base model, where the image features are extracted without fusing information from instruction features. Results in Tab.~\ref{tab:ablation study} show that adopting the editing transformer leads to \textbf{-0.6\%} mAP performance drop in the MTL scenario. Consistent results can be observed in STL, indicating the effectiveness of the instruction integrating design in the editing transformer.

\noindent \textbf{Adaptive Triplet Loss.}
Tab.~\ref{tab:ablation study} shows that adaptive triplet loss in MTL outperforms the traditional triplet loss by \textbf{+0.3\%} mAP on average, indicating that the proposed loss boosts the model to learn more discriminative features following different instructions. In the STL scenario, for CC-ReID and Trad-ReID, the instructions are fixed sentences leading to the same performance of adaptive/traditional triplet loss. However, in CTCC-ReID and LI-ReID, where instructions vary among samples, using adaptive triplet loss brings about \textbf{+0.7\%}, \textbf{+0.5\%} mAP gain, which shows the effectiveness of adaptive triplet loss in learning both identity and instruction similarity.

\begin{figure*}[t]
    \centering
    \includegraphics[width=\linewidth, height=0.185\linewidth]{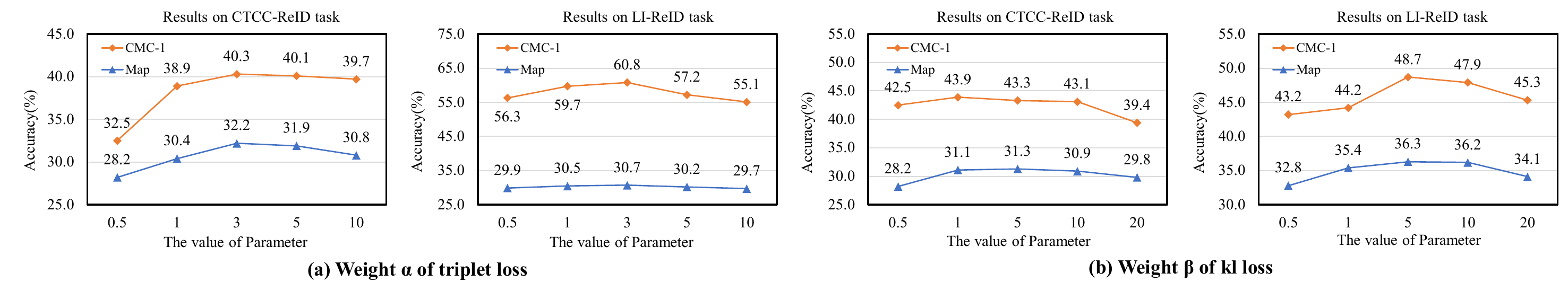}
    \caption{Ablation experiments on the CTCC-ReID and LI-ReID tasks to test the effects of different combinations of loss weights in the loss function.}
    \label{fig:parameter_ablation}
\end{figure*}

\noindent \textbf{Ablation studies on hyper-parameters.}
% We conducted ablation experiments on several important hyper-parameters used in the method and model training, with experimental results based on the CTCC-ReID and LI-ReID tasks, using mAP and rank1 evaluation metrics. Fig.~\ref{fig:parameter_ablation} (a) shows the impact of different learning rates on the experimental results of the two tasks under the STL training method. We increased the learning rate from 0.000001 to 0.0001, and the results of both metrics showed a trend of initially increasing and then stabilizing. The performance peaked when the learning rate was set to 0.00001 for both tasks, so we selected this learning rate as the default value for the STL training method in each ReID task experiment. 
As depicted in Fig.~\ref{fig:parameter_ablation} (a), we perform ablation experiments to analyze the effect of loss weights on the identity loss and adaptive triplet loss components within the total loss during the training of the IRM model. The best results are achieved when the triplet loss weight $\alpha$ is set to 3. Similarly, Fig.~\ref{fig:parameter_ablation} (b) illustrates the impact of different weights on the contrastive loss components within the total loss during the training of the IRM++ model. We vary the hard label contrastive loss weight coefficient from 0.5 to 20. Notably, when the weight $\beta$ is set to 5, except the rank-1 result of CTCC-ReID is marginally lower compared to the $\beta$ 1, all other experiments get the best results. Based on these experimental findings, we establish the default values of the triplet loss weight as 3 and the hard label contrastive loss weight as 5 for experiments across each ReID task.

\begin{figure*}[t]
    \centering
    \includegraphics[width=\linewidth]{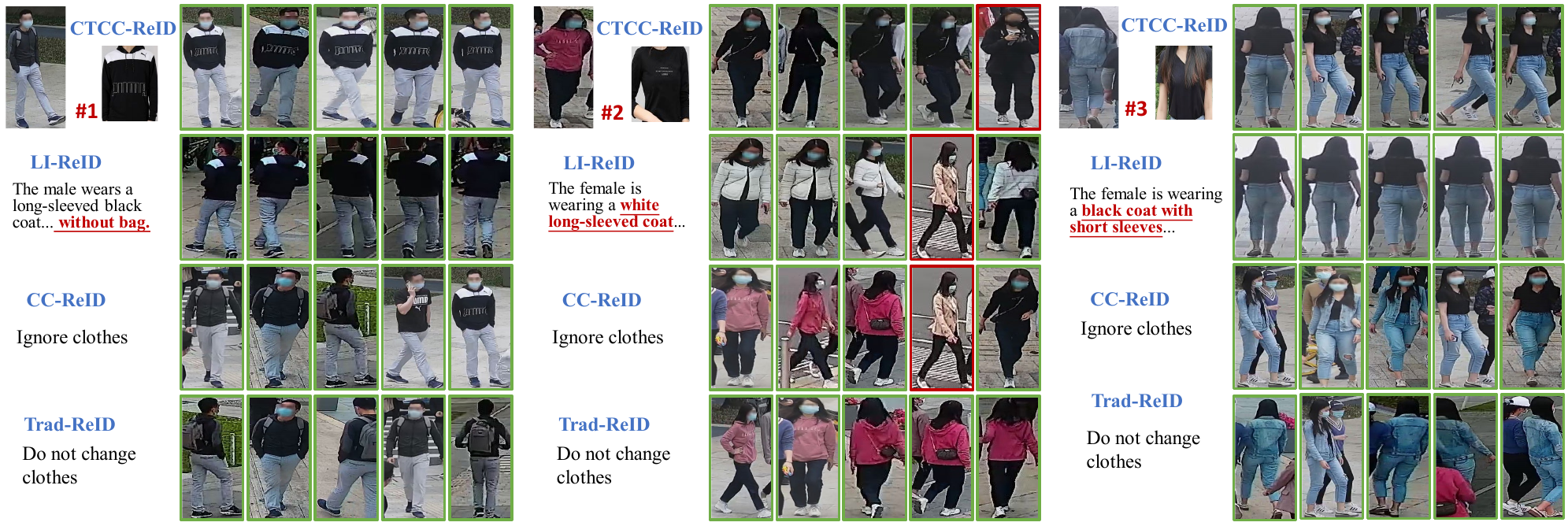}
    \caption{Illustration of all tasks retrieval results. We visualize the task-specific instructions on three people as examples. To clearly describe the retrieval results, we use green boxes to mark true retrieval samples and red boxes mean false matches.}
    \label{fig:vis}
\end{figure*}

\begin{figure*}[t]
    \centering
    \includegraphics[width=\linewidth]{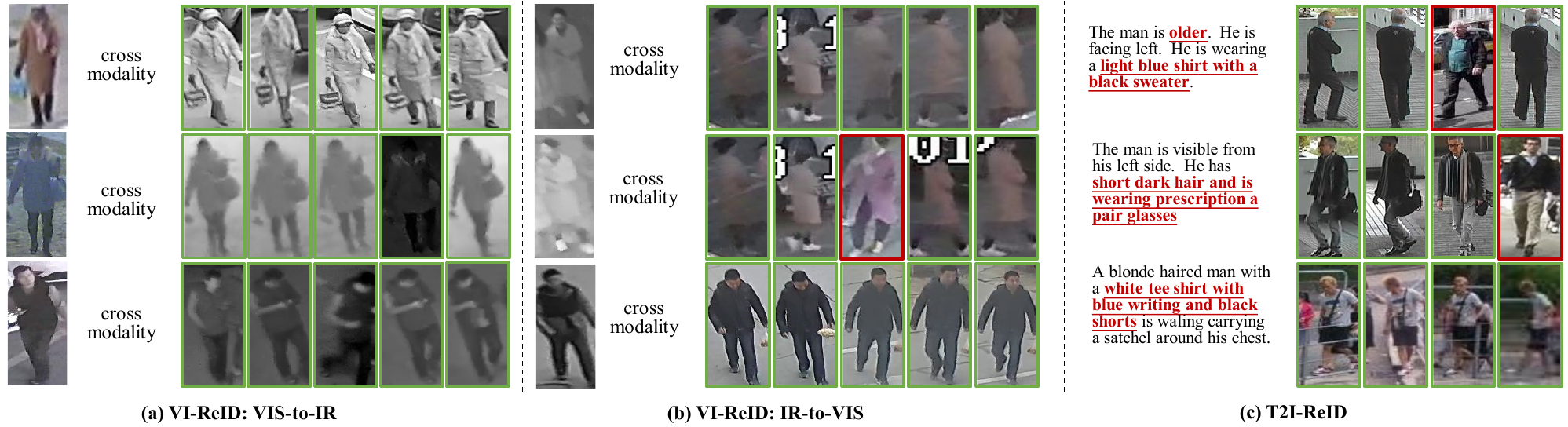}
    \caption{Illustration of VI-ReID and T2I-ReID tasks retrieval results. We visualize both VIS-to-IR and IR-to-VIS mode results on three people in the LLCM dataset. There are about four images for each person in  CUHK-PEDES, we visualize the top 4 results for T2I-ReID on the CUHK-PEDES dataset as examples. Green and red boxes mean true and false matches.}
    \label{fig:vis_vi}
\end{figure*}

\subsection{Visualization}

\noindent \textbf{Visualization of retrieval results.}
We visualize the retrieval results of CTCC-ReID, LI-ReID, CC-ReID and Trad-ReID tasks in Fig.~\ref{fig:vis} and VI-ReID, T2I-ReID in Fig.~\ref{fig:vis_vi}. Given a query image, IRM not only retrieves the right person from the gallery but also finds specific target images of the person following the instruction. Concretely, for CTCC-ReID, IRM retrieves images of query persons wearing instructed clothes as shown in the first row. For LI-ReID, IRM effectively parses information from languages such as bag condition (\emph{e.g.}, row 2 person {\#1}) and clothes attribute (\emph{e.g.}, row 2 person {\#2,3}) to find the correct image. For CC-ReID, with the ``Ignore clothes'' instruction, our method focuses on biometric features and successfully retrieves the person in the case of changing clothes, \emph{e.g.} the 4$^{th}$ and 5$^{th}$ images of row 3 person {\#1}, which the clothes are different from the query image. For Trad-ReID shown in row 4, ``do not change clothes'' instructs IRM to pay attention to clothes, a main feature of a person's image. In this case, IRM retrieves images with the same clothes in the query image. For failed cases, \emph{e.g.}, 5$^{th}$ image row 1 and 4$^{th}$ image row 2,3 of person {\#2}, the similar features like hairstyle, body shape, and posture between retrieved images and the query image cause the mismatch of IRM. For VI-ReID, we visualize the retrieval results for both VIS-to-IR and IR-to-VIS modes. IRM can retrieve the correct cross-modality images even in low visual conditions. For T2I-ReID, when there are images of different identities in the gallery but with representations consistent with the text description (\emph{e.g.}, row 1 3$^{rd}$ and row 2 person 4$^{th}$ images in \textcolor{black}{Fig~\ref{fig:vis_vi} (c)}), it may introduce some noise to IRM. However, IRM is still capable of indexing most of the correct results based on the given descriptions.

\begin{figure*}[t]
    \centering
    \includegraphics[width=\linewidth, height=0.2\linewidth]{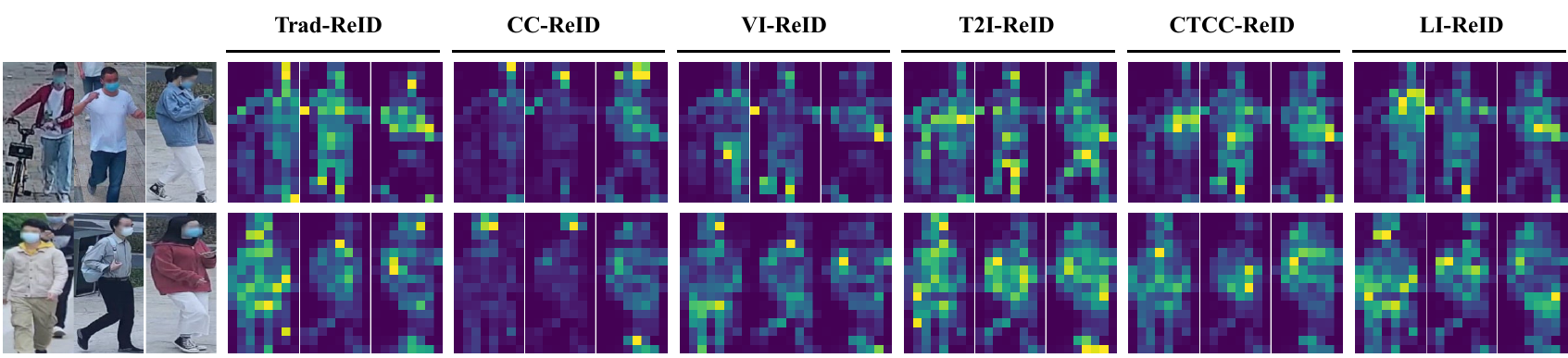}
    \caption{Illustration of the attention maps for instruct-ReID. We conduct inference for Trad-ReID, CC-ReID, VI-ReID, T2I-ReID, CTCC-ReID, and LI-ReID tasks on the same image using our proposed IRM model. Visualizing the attention maps helps us explore how well the model comprehends various tasks. }
    \label{fig:vis_attention}
\end{figure*}

\noindent \textbf{Visualization of attention maps.}
We visualize attention feature maps to understand what IRM has learned across the 6 ReID tasks. As illustrated in Fig.~\ref{fig:vis_attention}, we conduct inference for all 6 ReID settings on the same image, demonstrating that IRM learns to attend to different regions based on the instruction. The feature maps of Trad-ReID mainly emphasize holistic features such as the face and upper clothing, \emph{e.g.}, 1$^{st}$ to 3$^{rd}$ attention map images of row one. For CC-ReID, the feature maps focus on clothing-irrelevant features to mitigate interference from inconsistent clothes, \emph{e.g.}, 4$^{th}$ to 6$^{th}$ attention map images. In VI-ReID, the attention maps exhibit a balanced focus on human features, yet still capture regions with more prominent characteristics, \emph{e.g.}, the 1$^{st}$ attention map image of VI-ReID example row one. In T2I-ReID and LI-ReID, the introduction of specified language descriptions leads the model to highlight certain attributes, such as backpacks and trousers, \emph{e.g.}, 2$^{nd}$ attention map image of LI-ReID example row two. For CTCC-ReID, the model emphasizes the upper clothing, indicating potential clothes template changing. These visualizations of attention maps qualitatively demonstrate the efficacy of our method in addressing various ReID tasks.

\begin{figure*}
  \centering
  \includegraphics[width=\linewidth]
  {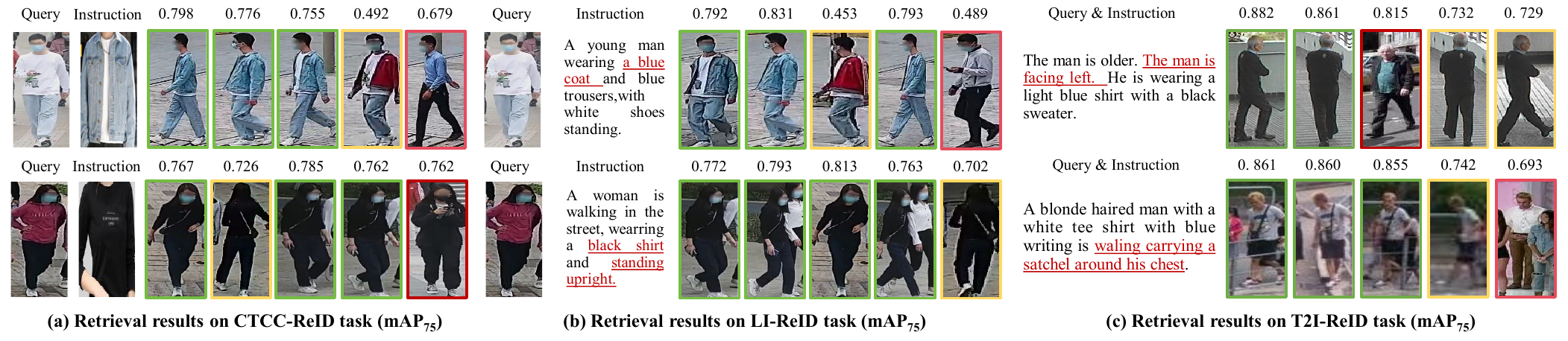}
  \caption{Visualization of retrieval results for CTCC-ReID, LI-ReID, and T2I-ReID using the mAP$\tau$ metric with a threshold set at 0.75. Positive samples are highlighted in green boxes, negative samples in red boxes, and retrieval results with correct identities but instruction similarity below the threshold, corrected to negative samples, are marked with yellow boxes.}
  \label{fig:vis_metric}
\end{figure*}

\noindent \textbf{Visualization of mAP$\tau$.}
In this subsection, we visually validate the improvement of the proposed evaluation metric over the traditional mAP metric. We visualize the indexing results on the CTCC-ReID, LI-ReID, and T2I-ReID tasks, setting the threshold to 0.75. The visualization demonstrates that mAP$\tau$ provides a more accurate measure of both identity correctness and retrieval description consistency in the retrieval results. The green box indicates correctly indexed positive samples, the red box indicates incorrectly indexed negative samples and the yellow box indicates indexed results with correct IDs but with instruction similarity below the set threshold, which are then relabeled as negative samples. For instance, in \textcolor{black}{Fig.~\ref{fig:vis_metric} (a)} for person 1, the 4$^{th}$ image, and in \textcolor{black}{Fig.~\ref{fig:vis_metric} (b)} for person 1, the 3$^{rd}$ image, although the model successfully indexed images with the same ID as the query image, it did not find the person wearing the blue jacket as required by the query instruction. In these scenarios, the proposed mAP$\tau$ evaluation metric can successfully filter out retrieval results that do not match the query instruction description. In addition to assessing whether clothing features match, this evaluation metric can also validate the ability of the ReID model to recognize fine-grained features such as body posture and accessories. For example, in \textcolor{black}{Fig.~\ref{fig:vis_metric} (b)} for person 2, the 5$^{th}$ image (body orientation does not match the retrieval target), and in \textcolor{black}{Fig.~\ref{fig:vis_metric} (c)} for person 2, the 5$^{th}$ image (unable to discern the status of the carried accessory). Setting the threshold for mAP$\tau$ is a crucial consideration, as seen in \textcolor{black}{Fig.~\ref{fig:vis_metric} (c)} for person 1, the 4$^{th}$ and 5$^{th}$ images, where an excessively large threshold may classify some true positive samples as negative samples. Choosing an appropriate threshold is a topic worthy of future research.

\section{Conclusion}
% \vspace{-0.5em}
This proposes one unified instruct-ReID task to jointly tackle existing traditional ReID, clothes-changing ReID, clothes template based clothes-changing ReID, language-instruct ReID, visual-infrared ReID, and text-to-image ReID tasks, which holds great potential in social surveillance. To tackle the instruct-ReID task, we build a large-scale and comprehensive OmniReID++ benchmark and a generic framework with an adaptive triplet loss. We hope our OmniReID++ can facilitate future works such as unified network structure design and multi-task learning methods on a broad variety of retrieval tasks.

{\small
\bibliographystyle{unsrt}
\bibliography{bib}
}
% \vspace{-3em}
\begin{IEEEbiography}[{\includegraphics[width=1in,height=1.25in,clip,keepaspectratio]{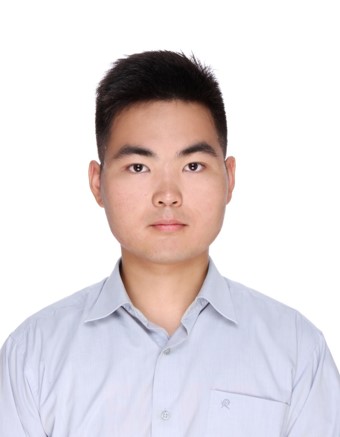}}]{Weizhen He} received the B.S.degree in College of Electrical Engineering from Zhejiang University, Hangzhou, China, in 2021. He is currently working toward the Ph.D. degree in the major of Control Theory and Control Engineering in Zhejiang University. His research interests include human-centric artificial intelligence, person re-identification and object detection.
% \vspace{-3em}
\end{IEEEbiography}

\begin{IEEEbiography}[{\includegraphics[width=1in,height=1.25in,clip,keepaspectratio]{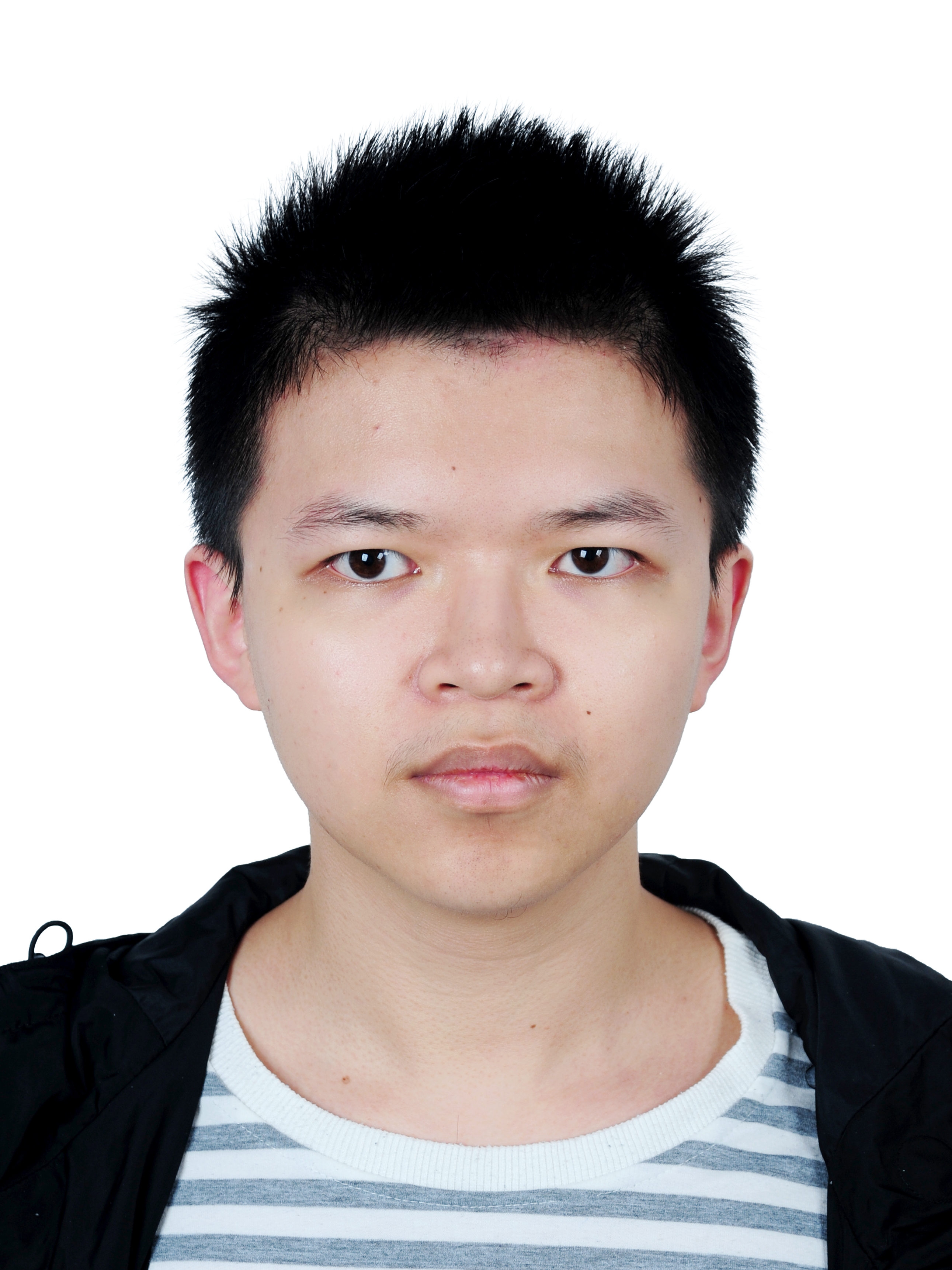}}]{Yiheng Deng} received the B.S.degree in School of electrical engineering from Chongqing University, Chongqing, China, in 2018. He is currently working toward a Ph.D. degree in the major of electrical engineering at Zhejiang University, Hangzhou, China. His research interests include deep learning, person re-identification and object detection.
% \vspace{-3em}
\end{IEEEbiography}

\begin{IEEEbiography}[{\includegraphics[width=1in,height=1.25in,clip,keepaspectratio]{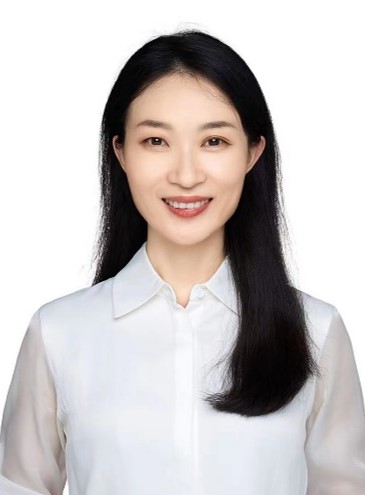}}]{Yunfeng Yan} received the Ph.D. degree in Electrical Engineering from Zhejiang University, Hangzhou, China, in 2019. She is currently an Associate Research Fellow with the College of Electrical Engineering, Zhejiang University, Hangzhou, China. Her research interests include computer vision, operation situational awareness, and abnormality monitoring of distributed generation equipment.
% \vspace{-4em}
\end{IEEEbiography}

\begin{IEEEbiography}[{\includegraphics[width=1in,height=1.25in,clip,keepaspectratio]{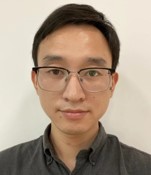}}]{Feng Zhu} is currently a Research Director in Smart City Group of SenseTime, leading a talented research development team with over 40 full-time researchers, engineers, and interns. His team supports SenseFoundry, a one-stop software platform tailored for Smart City management, delivering cutting-edge Computer Vision and Deep Learning techniques for large-scale applications in Smart City, such as City Safety, and Traffic Management. Feng ZHU received a B.E. degree and Ph.D. degree in Electronic Engineering in 2011 and 2017, respectively, from the Department of Electronic Engineering and Information Science, University of Science and Technology of China (USTC).
% \vspace{-2em}
\end{IEEEbiography}

\begin{IEEEbiography}[{\includegraphics[width=1in,height=1.25in,clip,keepaspectratio]{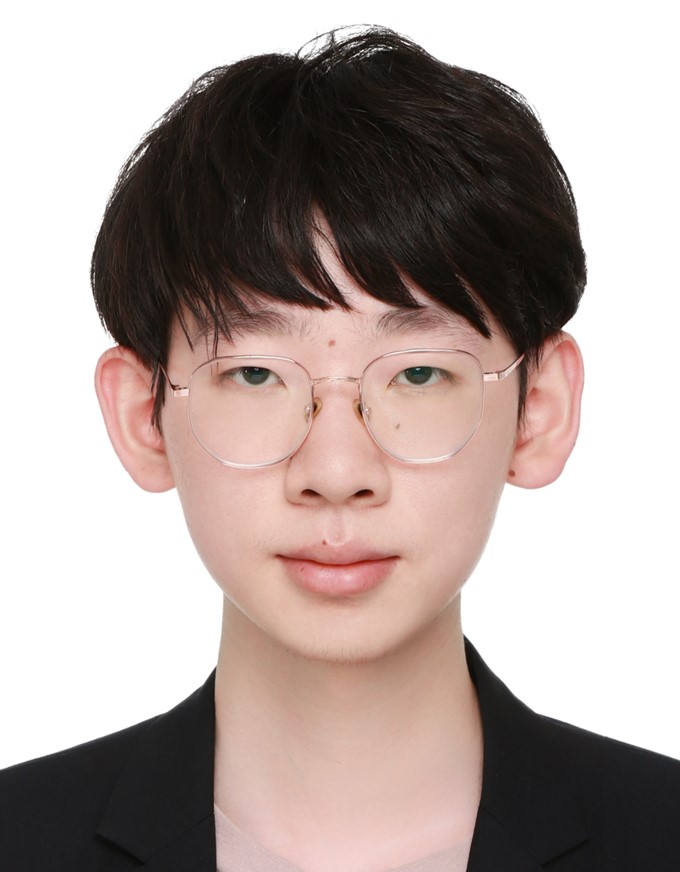}}]{Yizhou Wang}
is an assistant researcher at Shanghai AI Laboratory. He received Master's degree in Control Theory and Control Engineering from Zhejiang University in 2023 and Bachelor's degree in Automation from Zhejiang University in 2020. His research interests are concentrated in the areas of unsupervised learning, transfer learning, and the advancement of human-centric unified models in computer vision.
% \vspace{-7em}
\end{IEEEbiography}

\begin{IEEEbiography}[{\includegraphics[width=1in,height=1.25in,clip,keepaspectratio]{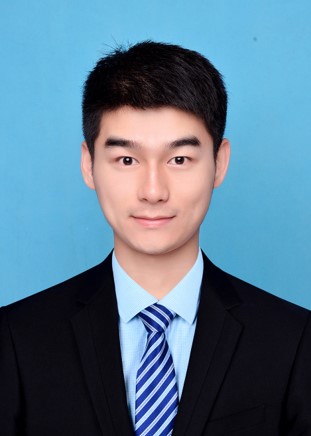}}]{Lei Bai} is a research scientist at Shanghai AI Laboratory. Prior to that, he was a Postdoctoral Research Fellow at the University of Sydney, Australia. He received his PhD degree in Computer Science from UNSW Sydney under the supervision of Prof. Lina Yao and Prof. Salil Kanhere. His research interests lay in Spatial-Temporal Generative Learning and it's applications in cross-discipline scenarios, such as climate and weather forecasting, and intelligent transportation. Dr. Bai has published more than 60 papers in top-tier conferences and journals, including Nature Communications, IEEE TPAMI, IEEE TIP, NeurIPS, CVPR, IJCAI, and KDD.
% \vspace{-7em}
\end{IEEEbiography}

\begin{IEEEbiography}[{\includegraphics[width=1in,height=1.25in,clip,keepaspectratio]{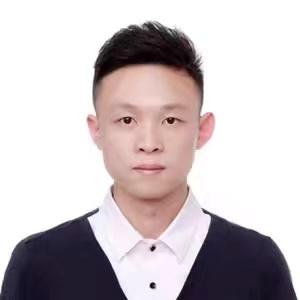}}]{Qingsong Xie} received the B.S. degree from the University of Electronic Science and Technology of China, Chengdu, China,  and PhD degree from Shanghai Jiao Tong University, Shanghai, China. His research interests include medical signal analysis, computer vision,  multi-modal understanding and generation.
% \vspace{-8em}
\end{IEEEbiography}

\begin{IEEEbiography}[{\includegraphics[width=1in,height=1.25in,clip,keepaspectratio]{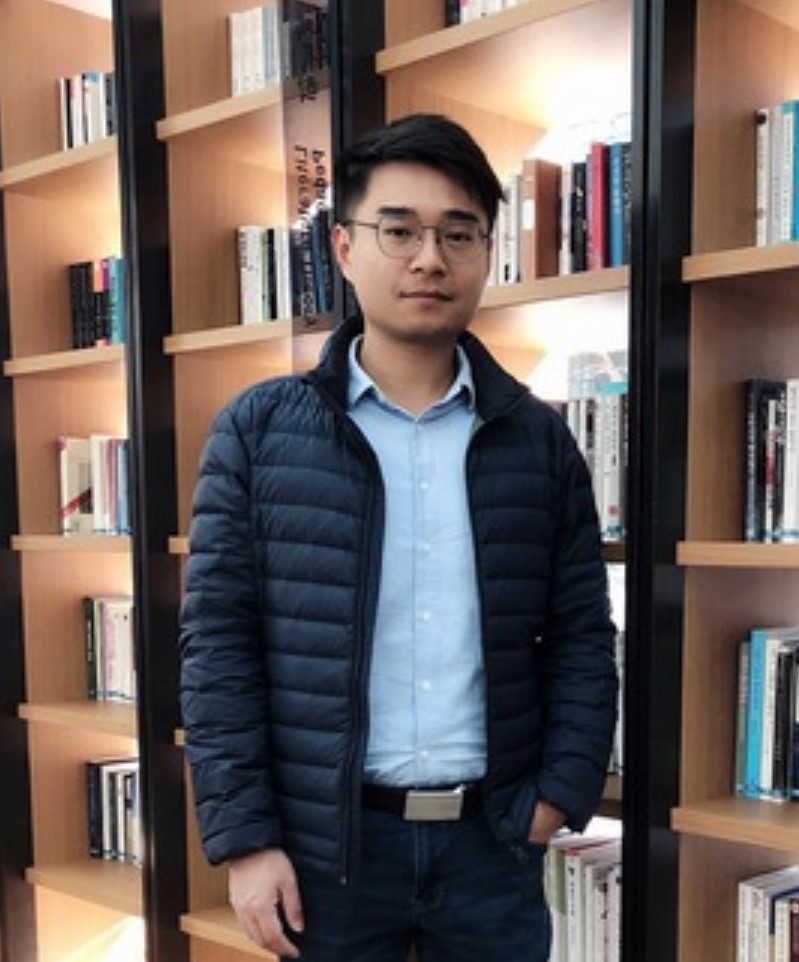}}]{Rui Zhao} is the Research Executive in the Smart
City Group of SenseTime. He is also the lead
researcher at Qing Yuan Research Institute of Shanghai Jiaotong University. He obtained Ph.D. from
the Chinese University of Hong Kong in 2015.
His research interest is computer vision and deep
learning. He has more than 100 publications, with
around 10,000 citations.
% \vspace{-7em}
\end{IEEEbiography}

\begin{IEEEbiography}[{\includegraphics[width=1in,height=1.25in,clip,keepaspectratio]{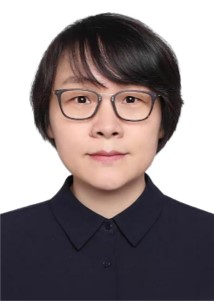}}]{Donglian Qi} received the Ph.D. degree from the School of Electrical Engineering, Zhejiang Univer- sity, China, in 2002. She is currently a Full Professor and a Ph.D. Advisor with Zhejiang University. Her recent research interest covers intelligent informa- tion processing, chaos systems, and nonlinear theory and application.
\vspace{-7em}
\end{IEEEbiography}

\begin{IEEEbiography}[{\includegraphics[width=1in,height=1.25in,clip,keepaspectratio]{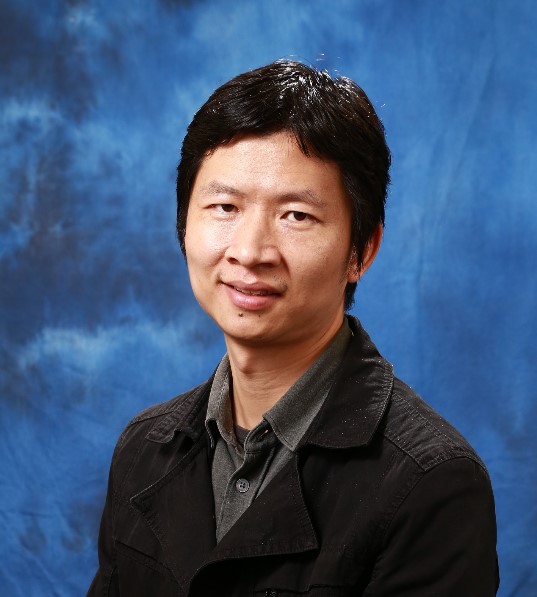}}]{Wanli Ouyang} (Senior Member, IEEE) received the Ph.D. degree from the Department of Electronic Engineering, The Chinese University of Hong Kong. His research interests include deep learning and its application to computer vision and pattern recognition, image, and video processing. He was awarded the Australian Research Council Future Fellowship, meaning that he will be exempted from teaching and can focus on research in the next four years.
\vspace{-6em}
\end{IEEEbiography}

\begin{IEEEbiography}[{\includegraphics[width=1in,height=1.25in,clip,keepaspectratio]{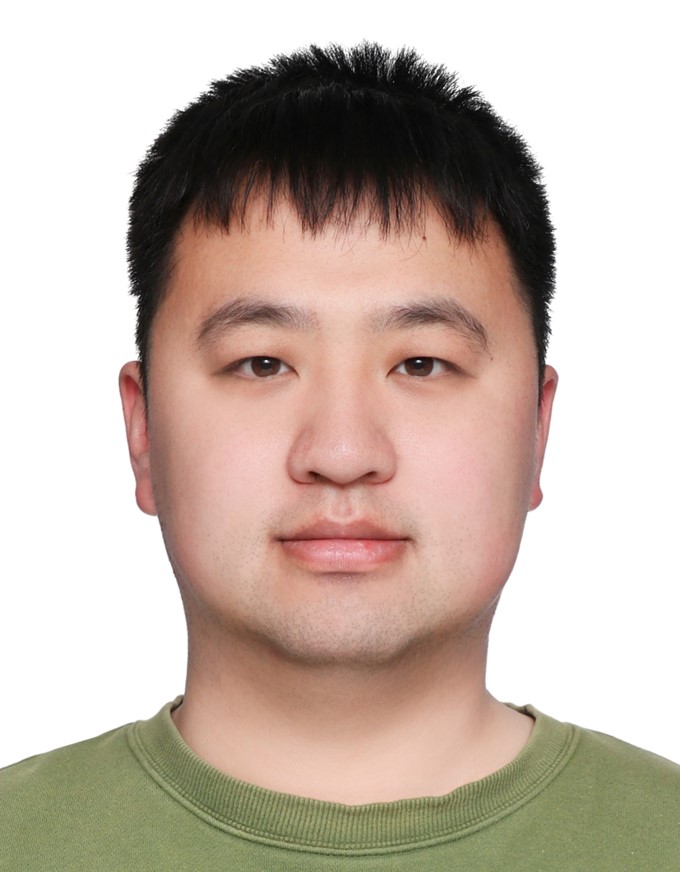}}]{Shixiang Tang} received the Ph.D degree from the University of Sydney. Prior to that, he received the Master of Philosophy from the Chinese University of Hong Kong in 2018 and Bachelor of Science from Fudan University. His interests lie in machine learning and computer vision, especially self-supervised learning and foundation models. He has published about 10 papers in top-tier conferences and journals, e.g., CVPR, ICCV, NeurIPS, Nature Physics and Nature Materials.
\vspace{-3em}
\end{IEEEbiography}

\newpage

 

% \end{IEEEbiography}



\vfill

\end{document}